\documentclass[10pt,twocolumn,letterpaper]{article}

\usepackage[pagenumbers]{cvpr} 

\usepackage[dvipsnames]{xcolor}
\usepackage{multirow}
\usepackage{colortbl}
\usepackage{algorithm2e}
\usepackage{xspace}
\usepackage{graphicx}
\usepackage{overpic}
\usepackage{wrapfig}

\definecolor{lightgray}{gray}{0.9}
\definecolor{mediumgray}{gray}{0.7}

\usepackage{amssymb}
\usepackage{pifont}
\newcommand{\cmark}{\ding{51}}%
\newcommand{\xmark}{\ding{55}}%

\newcommand{\dname}{\emph{CADTalk}\xspace}
\newcommand{\dnameC}{\emph{CADTalk-Cube}\xspace}
\newcommand{\dnameCH}{\emph{CADTalk-Cube$^H$}\xspace}
\newcommand{\dnameCL}{\emph{CADTalk-Cube$^L$}\xspace}
\newcommand{\dnameE}{\emph{CADTalk-Ellip}\xspace}
\newcommand{\dnameEH}{\emph{CADTalk-Ellip$^H$}\xspace}
\newcommand{\dnameEL}{\emph{CADTalk-Ellip$^L$}\xspace}
\newcommand{\dnameR}{\emph{CADTalk-Real}\xspace}

\newcommand{\methodName}{\emph{CADTalker}\xspace}

\newcommand{\revi}[1]{{\color{black}#1}}

\newcommand{\prag}[1]{{\noindent \textbf{#1}}}

\definecolor{cvprblue}{rgb}{0.21,0.49,0.74}
\usepackage[pagebackref,breaklinks,colorlinks,citecolor=cvprblue]{hyperref}

\title{CADTalk: An Algorithm and Benchmark for Semantic Commenting \\ of CAD Programs}

\author{
    Haocheng Yuan$^1$ \quad Jing Xu$^1$ \quad Hao Pan$^2$ \quad Adrien Bousseau$^{3,6}$  \quad Niloy J. Mitra$^{4,5}$ \quad Changjian Li$^1$ \vspace{2mm}\\
    $^1$University of Edinburgh \quad $^2$Microsoft Research Asia \quad $^3$Inria, Université Côte d'Azur \\
    $^4$University College London \quad $^5$Adobe Research \quad $^6$Delft University of Technology
}

\begin{document}

\twocolumn[{
\renewcommand\twocolumn[1][]{#1}
\maketitle 
\begin{center}
\vspace{-6mm}
    \centering
    \includegraphics[width=1.0\textwidth]{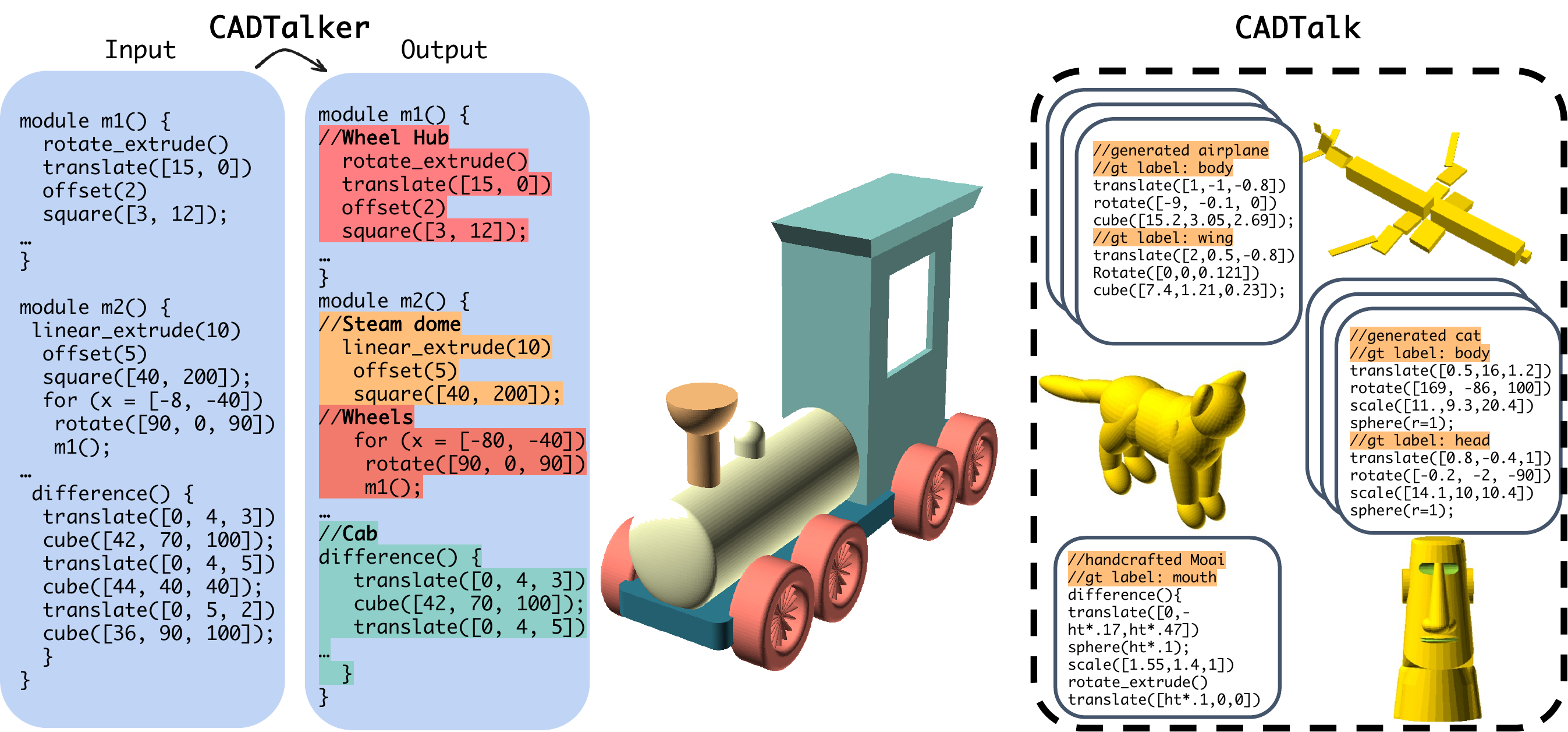}
    \captionof{figure}{Given a CAD program as input, our algorithm -- \methodName~ -- automatically generates comments before each code blocks to describe the shape part that is generated by the block (left). We evaluate our algorithm on a new dataset of commented CAD programs -- \dname~ -- that contains both human-made and machine-made CAD programs (right).}
    \label{fig:teaser}
\end{center}
}]

\begin{abstract}

CAD programs are a popular way to compactly encode shapes as a sequence of operations that are easy to parametrically modify. 
However, without sufficient semantic comments and structure, such programs can be challenging to understand, let alone modify. 
We introduce the problem of semantic commenting CAD programs,wherein the goal is to segment the input program into code blocks corresponding to semantically meaningful shape parts and assign a semantic label to each block.
We solve the problem by combining program parsing with visual-semantic analysis afforded by recent advances in foundational language and vision models. Specifically, by executing the input programs, we create shapes, which we use to generate conditional photorealistic images to make use of semantic annotators for such images. We then distill the information across the images and link back to the original programs to semantically comment on them. 
Additionally, we collected and annotated a benchmark dataset, \dname, consisting of 5,288 machine-made programs and 45 human-made programs with ground truth semantic comments. 
We extensively evaluated our approach, compared it to a GPT-based baseline, and an open-set shape segmentation baseline, and reported an $83.24\%$ accuracy on the new \dname dataset.
\revi{Code and data: \href{https://enigma-li.github.io/CADTalk/}{https://enigma-li.github.io/CADTalk/}.}

\end{abstract}    
\section{Introduction}
\label{sec:intro}

Computer-Aided-Design~(CAD) is the industry standard for representing 3D shapes as sequences of geometric instructions, also referred to as \emph{CAD programs}. 
These programs are compact, expressive, and provide a parametric way to edit shapes. Decades of research have focused on developing domain-specific CAD languages, tools for authoring and robustly executing CAD programs, and even automatically generating them. Popular frameworks include SketchUp, AutoCAD, FreeCAD, and Catia, to name a few. 

However, without semantic annotations or comments, CAD programs are challenging to parse and decipher as one has to mentally execute the programs to reveal semantic associations of code blocks with corresponding shape parts. 
This is particularly problematic when the one using the program differs from the one who authored it, being human or machine. 
We, as humans, are notorious for leaving sparse comments when writing CAD programs. Machines, on the other hand, can be made to leave systematic comments when generating programs, but these comments may not be semantically relevant nor follow a canonical structure. 
Without semantic comments and an associated structure, human users find it challenging to interpret and edit CAD programs. Even machines may struggle to learn from programs without a canonical structure and/or semantic association to the underlying shapes.  

In this paper, we introduce the problem of \textit{semantic commenting} of CAD programs. Other than the input program and the shape category it represents, we assume access to an execution module to transform any target program to its corresponding 3D shape. Our goal is to segment the program into multi-level code blocks and assign each block a comment indicating the shape part it represents (Fig.~\ref{fig:teaser}).

Solving this problem requires overcoming multiple challenges.
First, CAD programs, especially the ones designed by humans, contain highly structured constructs (\eg, subroutines, control flows) that organize shape primitives recursively into meaningful parts, and their commenting demands structure parsing in the first place.
Second, working directly in the program domain prevents visual recognition of the corresponding shape parts. While executing the program produces a shape that can be rendered, CAD programs lack any material texture and, at best, have pseudo face colors -- executing them results in textureless projected images that are challenging to interpret by vision models trained on photographs. 
Third, CAD programs encode a vast array of shapes across arbitrary categories. Managing this diversity requires an approach that can generalize beyond a pre-defined set of semantic labels.

We introduce \methodName for semantic commenting of CAD programs (Fig.~\ref{fig:teaser}), which we achieve by combining program parsing with visual-semantic analysis afforded by recent advances in large language models~(LLMs) and foundational vision models. 
We first handle the nested program structures by performing a syntax tree analysis that identifies commentable code blocks at multiple levels of program constructs.
Then, we address the second challenge by leveraging conditional image generation to translate textureless CAD renderings into photorealistic images, which vision models can handle. We employ this photorealistic image synthesizer to render the CAD program from multiple views, resulting in a rich visual depiction of the shape produced by the program. 
We address the third challenge by using LLMs and vision models to segment and annotate the images with open-vocabulary labels. 
Finally, we aggregate the segmentation and part labels from the multiview images and transfer them back to relevant code blocks.

To evaluate our method and to foster future research on this problem, we also created {\dname} -- a new benchmark for semantic commenting of CAD programs. We collected 5300+ programs from a variety of data sources (online repositories \cite{baumann2018thingiverse,Runeman}, and shape abstraction algorithms \cite{liu2023marching,sun2019learning}), comprising human-designed and machine-generated CAD programs. 
We semi-automatically annotated these programs to provide ground truth comments.
Finally, we propose evaluation metrics to score any semantic commenting approach.
Based on this dataset, we have conducted statistical evaluation, a comprehensive ablation study inspecting several core components of our method, and comparisons with a GPT-based approach and an open-vocabulary shape segmentation method (i.e., PartSLIP~\cite{liu2023partslip}). All the evaluations demonstrate that \methodName sets a good baseline for future research on this problem.

In summary, this paper introduces the new task of semantic commenting CAD programs and the first benchmark and algorithm dedicated to this task. 

\section{Related Work}
\label{ses:rw}
\noindent \textbf{CAD programs.}
CAD programs represent 3D shapes as sequences of geometric instructions, which brings advantages in terms of compactness, editability, and modularity. However, CAD programs are notoriously difficult to author, as the effect of each instruction on the resulting shape can be difficult to foresee. Recent work in interactive modeling eases CAD authoring thanks to differentiable execution \cite{Cascaval2022,Michel2021}, which allows editing program parameters via direct shape manipulation, but does not allow modification of the program structure. Closer to our goal is the recent work by Kodnongbua et al.~\cite{Kodnongbua23}, who use large pre-trained image and language models to discover semantically meaningful parameter ranges from multi-view renderings of a CAD program. Using similar ingredients, we target the complementary task of generating semantic comments for code blocks that might include several parametric instructions. 

\begin{figure*}[!t]
    \centering
    \includegraphics[width=1.0\linewidth]{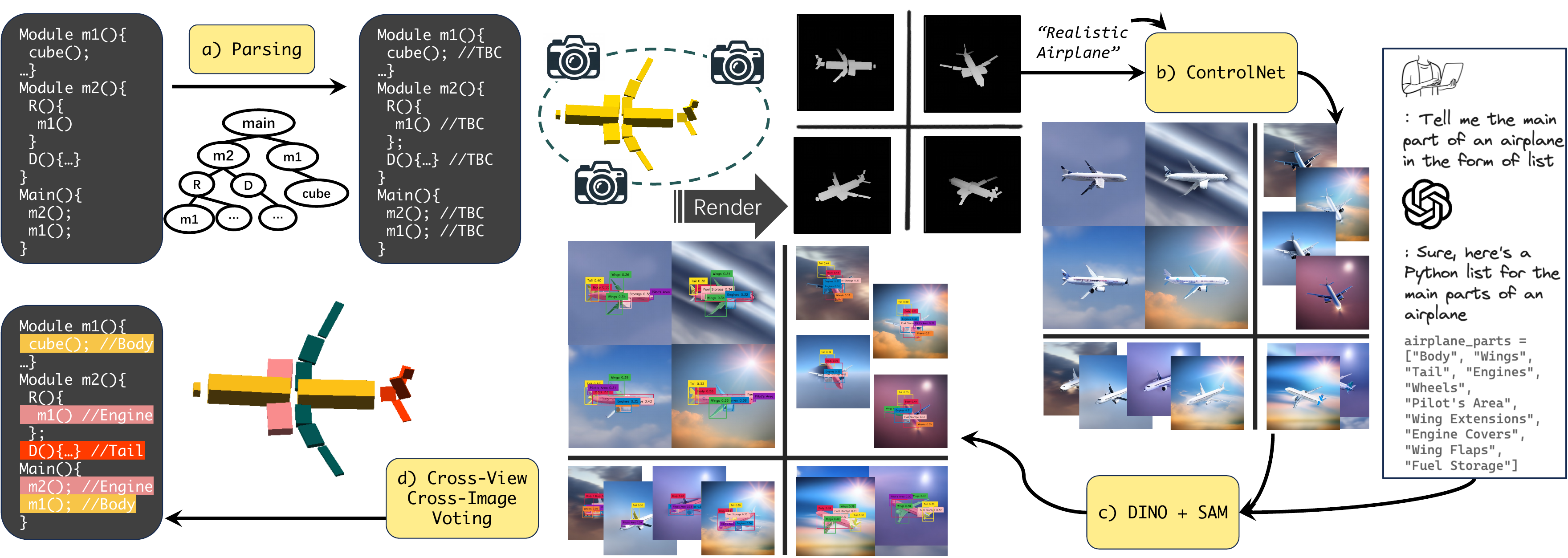}
    \vspace{-6mm}
    \caption{\textbf{Algorithm overview.} We first parse the input program to identify commentable code blocks, marked with TBC (a). We then execute the program and render the resulting shape under several viewpoints to obtain multiview depth maps, which we convert into realistic images using image-to-image translation (b). In addition, we obtain a list of part names of the shape from ChatGPT. We use these labels to segment semantic parts in the images using computer vision foundation models (c). Finally, we aggregate this semantic information across views by linking it to the code blocks that correspond to the segmented parts (d).}
    \vspace{-4mm}
    \label{fig:pipeline}
\end{figure*}

Another stream of research aims at automatically generating CAD programs
for reverse engineering \cite{du2018inversecsg,willis2020fusion,Yu_2022_CVPR}, sketch-based modeling \cite{Li:2022:Free2CAD}, or generative modeling \cite{jones2020shapeassembly,Wu_2021_ICCV,Xu2023}, see \cite{Ritchie2023} for a recent survey. 
While algorithms exist to structure the raw code generated by such methods, for instance by identifying compact macros that encapsulate repetitive parts \cite{Nandi2020,jones2023shapecoder}, in the absence of comments, machine-made code is difficult to interpret and build upon. We designed our benchmark dataset to include both human-made and machine-made CAD programs to foster research.

\prag{Program summarization.}
There is a large body of work on the automatic generation of comments for generic programming languages (C/C++, Python, Java, etc.). While early methods were based on template matching and text retrieval  \cite{Haiduc2010}, the field then adopted sequence-to-sequence neural models \cite{alon2018codeseq,iyer-etal-2016-summarizing} and most recently LLMs \cite{chen2021evaluating}. Such models are typically trained on large corpus of commented code, collected from code sharing platforms like StackOverflow \cite{iyer-etal-2016-summarizing} and GitHub \cite{chen2021evaluating}. Unfortunately, we are not aware of any large dataset of commented CAD programs that could be used to train such language-specific models. Besides, sequence-to-sequence models reason on raw code snippets rather than on their execution, and as such are not equipped to analyze the visual output of CAD programs. In contrast, we leverage CAD execution and rendering to convert the problem of code commenting into the problem of semantic image segmentation, allowing us to build upon pre-trained image models, and making our method agnostic to the input CAD language.
Nevertheless, we experimented with few-shot training strategies of an LLM for code commenting \cite{Brown2020,Ahmed2023}, and observed a surprising ability of the LLM to capture the geometric structure of an object category from only one example CAD program, although it struggles to generalize to different categories.

\prag{Semantic segmentation of images and shapes.}
We formulate commenting of CAD programs as a semantic segmentation task, making our approach related to prior work on shape segmentation and labeling. Recent approaches to label parts of 3D shapes employ deep learning, using neural networks tailored to voxel grids \cite{Wang2017}, 3D meshes \cite{Hanocka2019,dutt2024diffusion}, point clouds \cite{Qi_2017_CVPR}. Closest to our approach is the work of Kalogerakis et al.~\cite{Kalogerakis2017,MvDeCor2022}, who render the 3D shape from multiple views to leverage 2D CNNs for segmentation tasks. We use pre-trained \emph{open-vocabulary} object detection \cite{liu2023grounding} and segmentation \cite{kirillov2023segment}. Open-vocabulary methods rely on large image-language models \cite{CLIP2021} to recognize arbitrary objects in images \cite{liang2023open}, avoiding the need for a pre-defined set of labels. Our approach also relates to the recent PartSLIP \cite{liu2023partslip}, which combines multi-view rendering and a pre-trained image-language model GLIP \cite{li2021grounded} to segment 3D scans. A key difference is that the CAD programs we target are not equipped with realistic colors and textures, preventing the direct application of image models trained on photographs. 
We tackle this by converting our synthetic renderings into realistic images using image-to-image translation \cite{zhang2023ControlNet}. \revi{This strategy allows our approach to outperform a \emph{concurrent} zero-shot mesh segmentation method \cite{abdelreheem2023satr} (see comparison in the supplementary).}

\section{Commenting Programs with \methodName}
\label{sec:method}

Given a CAD program as input, our goal is to automatically insert comments around CAD instructions, such that these comments describe the semantic parts of the shape that the CAD instructions produce upon execution. For example, in Figure~\ref{fig:teaser}, the comments specify instructions responsible for the wheels and cab of a train.

Recognizing object parts from CAD instructions is a very difficult task, for both humans and machines, because code instructions only describe shapes indirectly, via geometric functions. Our key idea is to move the problem away from the program domain, towards the image domain where visual inference is much easier. To do so, we leverage CAD execution to produce the 3D shape corresponding to the program, we perform multiview rendering to obtain representative images of this shape, and we run computer vision models on these images to identify semantic parts and their labels. We then propagate the labels in the opposite direction, from images to the original CAD instructions. Fig.~\ref{fig:pipeline} illustrates the main steps of this cross-domain process.
We next describe how we leverage off-the-shelf foundation models to achieve these tasks.

While there exist many different CAD domain-specific-languages, and the algorithm we propose is agnostic to the chosen language, we built our prototype implementation around OpenSCAD \cite{OpenSCAD} -- a free CAD software based on Constructive Solid Geometry (CSG).

\subsection{Realistic Multiview Rendering}
\label{subsec:img_gen}
At the core of our approach is the idea of leveraging computer vision models trained on photographs to recognize semantic parts in CAD programs. To do so, our first step is to execute the program to produce a 3D shape, and render images of that shape from ten representative viewpoints. We distribute these viewpoints along a ring slightly above the shape, as shown in Fig.~\ref{fig:pipeline}.

However, many CAD programs only describe the geometry of a shape and do not include realistic textures and materials. Furthermore, some of the CAD programs we consider represent shapes in a simplified manner, missing geometric details or exhibiting spurious gaps between parts.
As a result, the images we obtain by directly rendering the program output do not look realistic, and as such are not well recognized by computer vision models (see the ablation study in Sec.~\ref{subsec:abl_study}). We tackle this challenge by translating these renderings into photorealistic images using ControlNet \cite{zhang2023ControlNet} -- a method that adds image conditioning to a large text-to-image diffusion model. Specifically, we render a depth map of the shape from each of the viewpoints, and we instruct ControlNet to turn each depth map into a realistic image of the shape, providing the category name of the object as a complementary text prompt. 

In contrast to related work that seeks to imbue 3D shapes with realistic texture maps \cite{Cao_2023_ICCV,Richardson2023}, our application scenario does not necessitate different views to share a consistent appearance. On the contrary, we observed that the subsequent step of recognizing object parts in the resulting images benefits from diversity in appearance, since object parts that might be ambiguous under a specific appearance might become recognizable under a different appearance. We build on this insight to further increase diversity by running ControlNet with four different seeds on each of the ten views, resulting in 40 realistic images of the shape in total.
We also note that for highly abstract shapes, the quality of the image-to-image translation increases when we process the depth maps with a morphologic closing~\cite{haralick1987image} to fill in small gaps (see supplemental materials for parameter settings). 
Fig.~\ref{fig:image_process} illustrates the realistic images we obtain.

\begin{figure}[!tb]
    \centering
    \includegraphics[width=0.9\linewidth]{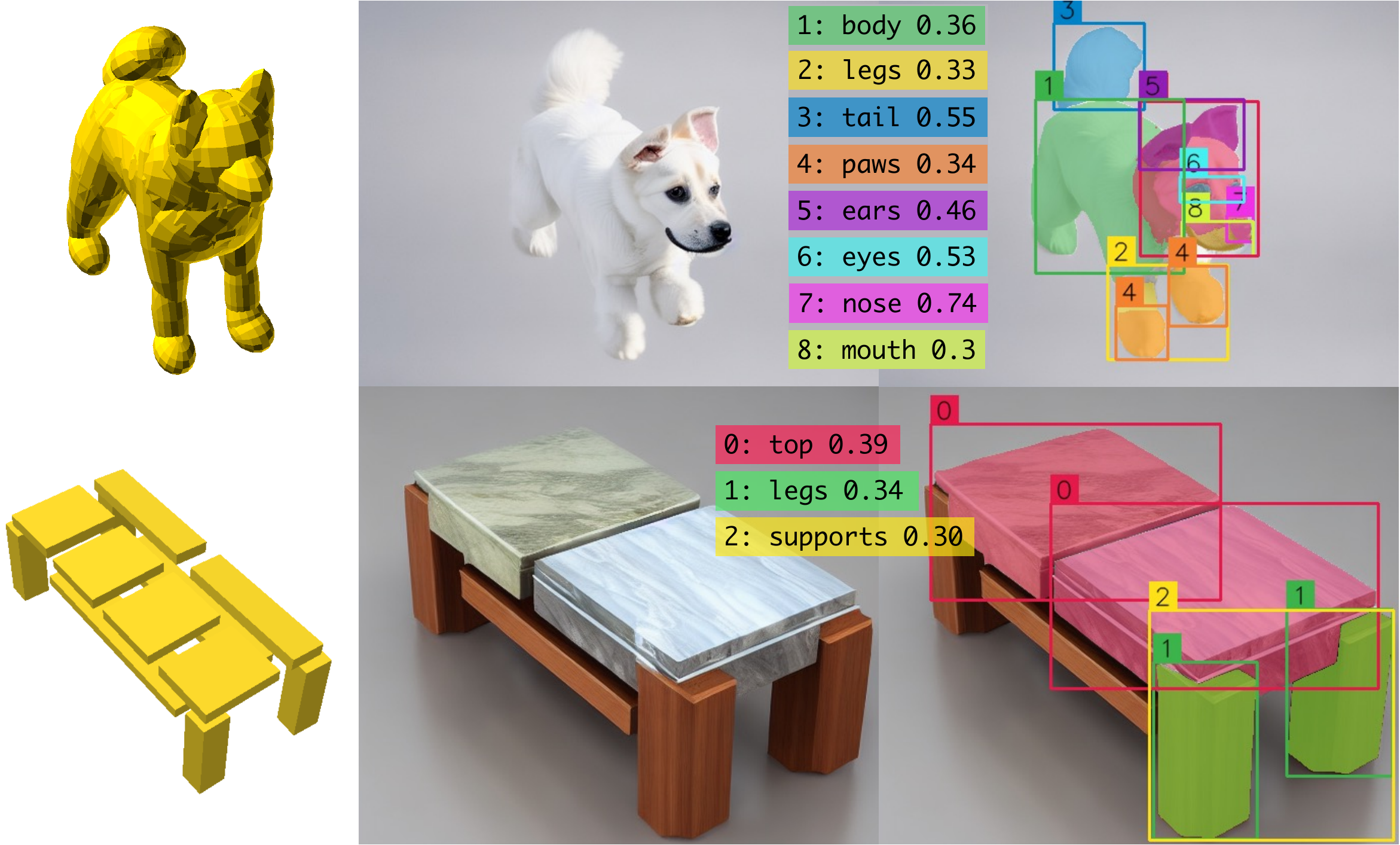}
    \vspace{-3mm}
    \caption{Given shapes~(left) after executing programs, we use ControlNet~\cite{zhang2023ControlNet} to convert rendered depth maps into realistic images (middle), which form a valid input for detection and segmentation models trained on photographs \cite{liu2023grounding,kirillov2023segment} (right).
    }
    \vspace{-4.5mm}
    \label{fig:image_process}
\end{figure}

\subsection{Part Detection and Segmentation}
\label{subsec:det_seg}
Given realistic images of the shape, our next step is to segment each image into semantically meaningful parts. We do so by leveraging both language and image foundation models.
Since we do not want to restrict ourselves to a fixed list of part labels, we turn to recent open vocabulary detection algorithms to identify relevant parts. Specifically, we use Grounding DINO~\cite{liu2023grounding}, a method that takes as input an image and the name of an object part of interest, and outputs a bounding box of that part, if present in the image. While users of our system can provide the list of parts to be detected, we found that ChatGPT-v4~\cite{ChatGPT:23} provides good suggestions of parts given the name of an object category.
We then convert each bounding box into a pixel-wise segment by feeding the image and the bounding box to the Segment Anything Model (SAM)~\cite{kirillov2023segment}. We denote $S^i_l$ the segment predicted for a given label $l$ on a given image $i$.
Fig.~\ref{fig:image_process} shows the segments and labels predicted for representative images.

\subsection{Part Label Voting}
\label{subsec:label_voting}
The last step of our algorithm consists of aggregating the semantic information of all 40 images of the shape and transferring this information to the corresponding code blocks.

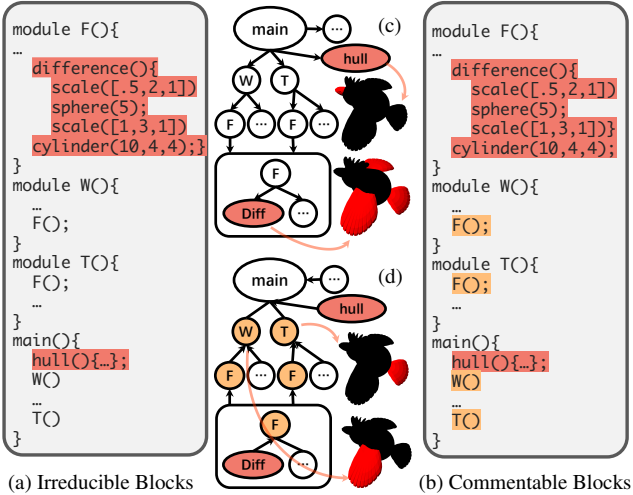
\begin{figure}[!tb]
    \centering
    \begin{overpic}[width=\linewidth]{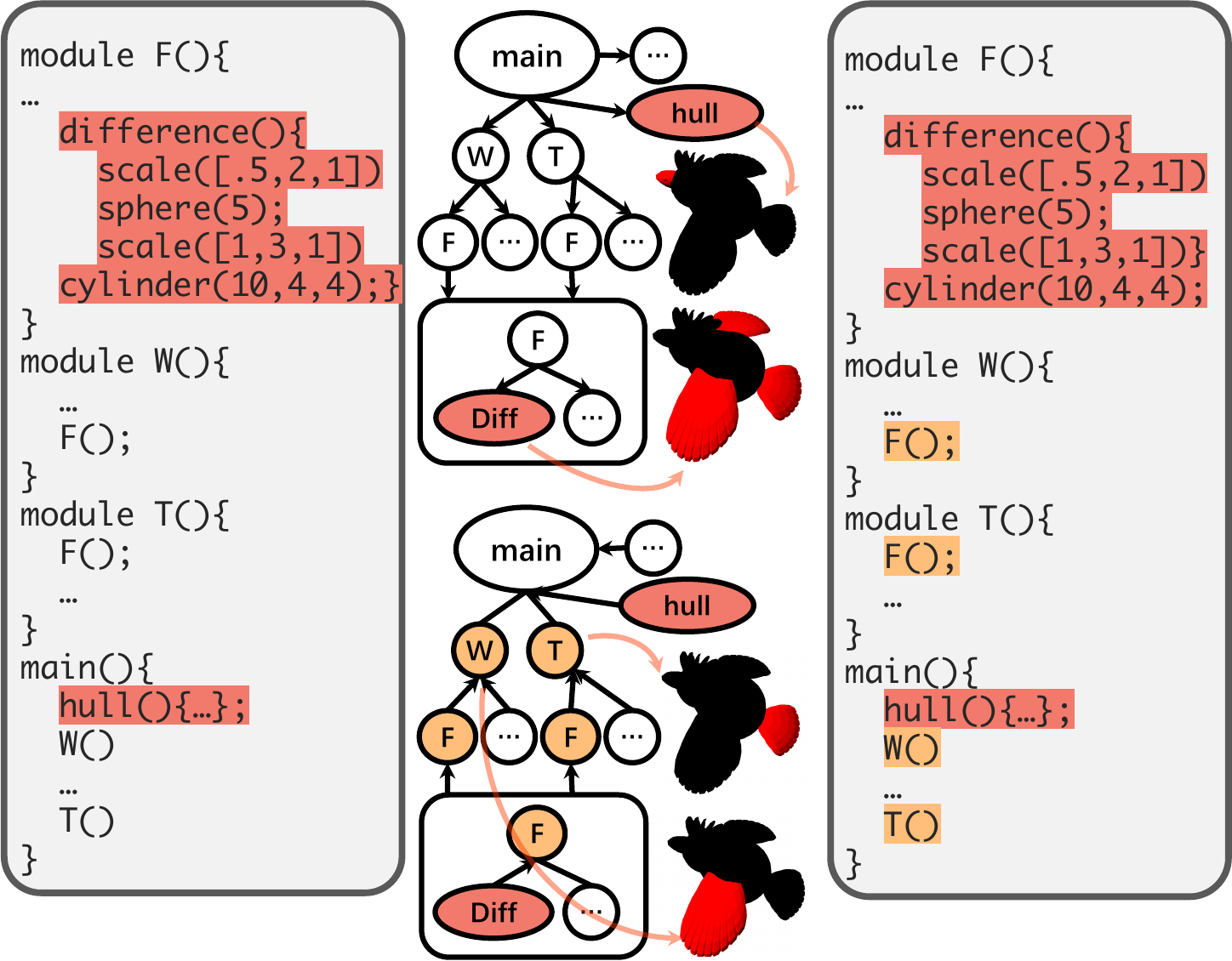} 
        \put(1, 0) {\footnotesize (a) Irreducible Blocks}
        \put(66.5, 0) {\footnotesize (b) Commentable Blocks}
        \put(60, 73) {\footnotesize (c)}
        \put(60, 33) {\footnotesize (d)}
    \end{overpic}
    \vspace{-6mm}
    \caption{\textbf{Program parsing.} Irreducible blocks are basic-level geometric primitives and their direct compositions (a), while commentable blocks are code blocks of different compositional levels that correspond to semantic comments (b). The downward traversal of the syntax tree is used to identify irreducible blocks (c) and the upward traversal to collect commentable blocks (d). Exemplar masks of commentable blocks are shown in (c) and (d) in red.}
    \label{fig:block_proc}
    \vspace{-5mm}
\end{figure} 

\prag{Program Parsing.}
We first parse the input program to identify all \emph{commentable} code blocks. Specifically, we construct the syntax tree of the input program (Fig.~\ref{fig:block_proc}c) and traverse the tree from the top downward using breadth-first search until we reach an \textit{irreducible} block, \emph{i.e.,} a sequence of instructions that corresponds either to a single geometric primitive (cuboid, ellipsoid, etc.) or to a difference,  intersection, or hull operation on primitives (Fig.~\ref{fig:block_proc}a). 
We label the irreducible blocks as commentable leaf nodes and traverse the tree upward to collect all commentable blocks as 
nodes parent of commentable nodes (Fig.~\ref{fig:block_proc}b,d).

\prag{Mapping code blocks to image pixels.}
For each block to be commented on, we render 10 views of the shape using a white color for the block of interest and a background color for all other blocks (see Fig.~\ref{fig:block_proc} (c)(d), where we rendered the blocks in red for visualization purpose). This procedure results in a set of binary masks $\{M^v_b\}$ that indicate for each view $v$ the visible pixels of block $b$, providing us with an explicit mapping between image pixels and code blocks. Fig.~\ref{fig:pipeline} illustrates the view setup, detailed view angles can be found in the supplemental material.

\prag{Aggregating semantic labels.}
We represent all possible label assignments via a matrix $C$, where each entry $C(b,l)$ quantifies the confidence of block $b$ to be assigned label $l$. We fill in this matrix by accumulating labeling confidence over all 40 images of the shape, in three steps.
In a first step we compute, for each image $i$ generated from view $v$, the confidence of label $l$ to be assigned to block $b$ as:
\begin{equation}
C^i(b,l) = C_{DINO}(i,l) \times IoU(M^v_b, S^i_l),
\label{eq:voting}
\end{equation}
where $C_{DINO}(i,l)$ is the confidence of label $l$ in image $i$ provided by Grounding DINO, $S^i_l$ is the segmentation mask provided by SAM, $M^v_b$ is the binary mask rendered for block $b$ in view $v$, and $IoU$ is the Intersection-over-Union. 
In a second step, we sum the confidence scores obtained for all 4 images of each view. Finally, in a third step, we sum the confidence scores over all 10 views.
We perform this aggregation in three steps to introduce intermediate filtering of poor labels. In practice, after each step, we set to zero the confidence of any label having a confidence below a threshold.
Finally, we assign to each block $b$ the label that received the highest cumulative confidence, $\max_{l}(C(b,l))$. 
\section{Building the \dname Dataset}
\label{sec:dataset}

While a few datasets of CAD programs have been recently introduced \cite{willis2020fusion,Wu_2021_ICCV}, these programs do not include ground-truth semantic comments. We address this issue by introducing \dname, a dataset of OpenSCAD \cite{OpenSCAD} programs enriched with part-based semantic comments (Figs.~\ref{fig:teaser} and \ref{fig:vis_main}). 
We first describe how we collected and commented on the programs in our dataset, and then introduce the metrics we used to evaluate the quality of automatically generated comments against this benchmark.

\subsection{Collecting CAD Programs}
\label{subsec:collect_prog}
We consider two distinct sources of programs to comment on, each raising its specific challenges. On the one hand, \emph{human-made programs} are often well-structured, with meaningful parts represented by independent code blocks. However, these programs exhibit a large diversity of instructions and program constructs, including macros, nested loops, and boolean operations to create intricate geometry, which can be difficult to reason about in the program domain.
On the other hand, \emph{machine-made} programs tend to obey a simplified CAD language with few instructions \cite{jones2020shapeassembly}, resulting in a rather flat structure without obvious meaningful code blocks. Moreover, such programs often generate abstract geometry made of simple primitives (cuboids, ellipsoids), which can be difficult to reason about in the visual domain. We built our dataset to contain representative programs of both sources.

\prag{Human-made programs.}
We gathered a set of 45 OpenSCAD programs from online repositories \cite{baumann2018thingiverse,Runeman}, or made by one of the authors.
For each such program, we first identify each \emph{commentable} code block as described in Sec.~\ref{subsec:label_voting} (see Fig.~\ref{fig:block_proc}).
We then manually comment on each such block with a label indicating the semantic part of the shape that the code block corresponds to.
When several irreducible blocks correspond to the same semantic part, we comment on each of them with the same label.
We used ChatGPT-v4~\cite{ChatGPT:23} 
to obtain a list of semantic part labels given the category name of the object of interest.

\begin{figure*}[!htb]
    \centering
    \includegraphics[width=1.0\linewidth]{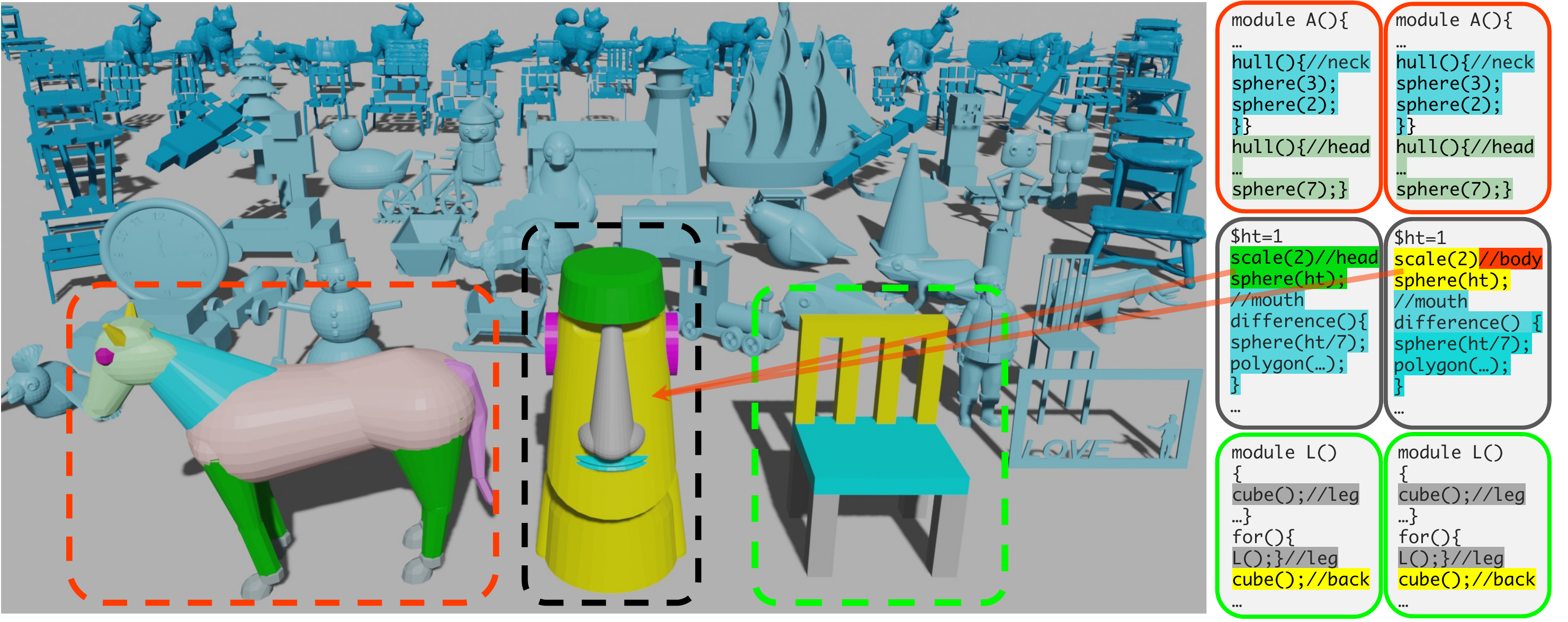}
    \vspace{-6mm}
    \caption{\textbf{\dname Dataset.}  Example shapes from \dname (left) along with ground-truth (right) and predicted comments (far right). In these examples, our prediction matches the ground truth, except for the Moai sculpture where \methodName labeled the ``head'' code block as ``body''. Machine-made shapes are rendered with dark blue and placed behind the human-made shapes rendered with light blue.}
    \vspace{-2mm}
    \label{fig:vis_main}
\end{figure*}

We name the resulting set of semantically commented programs \dnameR.
While this set is small due to the difficulty of finding real programs and commenting on them by hand, Figs.~\ref{fig:teaser} and \ref{fig:vis_main} show that it includes diverse human-made and organic shapes of varying complexity.

\prag{Machine-made programs.}
\label{subsec:gen_programs}Recent developments in shape analysis and program synthesis have enabled significant progress in generative models of CAD programs \cite{Ritchie2023}. While the most recent models seek to capture high-level geometric structures \cite{jones2021shapemod,jones2023shapecoder}, others produce a flat list of geometric primitives \cite{sun2019learning,jones2020shapeassembly}. Since the human-made programs we have collected already exhibit complex structures, we focused the second track of our dataset on flat programs.

We rely on automatic methods that convert 3D shapes into cuboid \cite{sun2019learning} and ellipsoid \cite{liu2023marching} abstractions.
By feeding these methods with semantically-labeled shapes from PartNet~\cite{mo2019partnet}, we obtain programs formed as unions of cuboids or ellipsoids, each such primitive forming a code block associated with a semantic label.
In the eventuality where a primitive covers several semantic parts of a shape, we associate that primitive with all the corresponding labels, each being treated as ground truth in our evaluation (i.e., either of the labels predicted is counted as correct).
Consecutive blocks with the same labels can then be naturally grouped.

We followed this procedure to generate programs for four object categories (Airplane, Chair, Table, and Animal). As illustrated in Fig.~\ref{fig:teaser}, the cuboid abstractions typically only contain a few primitives, resulting in a coarse approximation of the shapes, while the ellipsoid abstractions better reproduce curved surfaces and details.

To further evaluate the impact of shape approximation, we provide for each of the two types of abstraction (cuboids and ellipsoids) two levels of details (low and high), resulting in four sets of program named \dnameCL, \dnameCH, \dnameEL, and \dnameEH, respectively. On the one hand, a high level of details results in long programs to comment on, where multiple primitives need to be grouped under the same part label. On the other hand, a low level of details results in approximate shapes that are harder to recognize (Fig.~\ref{fig:abs_level} in the supplementary). 

Tab.~\ref{tab:dataset_stats} provides statistics of our dataset (number of programs, number of lines of code per program, number of semantic parts per program). In total, the dataset contains over 5300 commented CAD programs. The \dnameC and \dnameE tracks allow to evaluate performance of commenting algorithms at a large scale, while the \emph{CADTalk-Real} track provides a smaller albeit more diverse test set.

\begin{table}[!t]
    \centering
    \vspace{-4mm}
    \caption{\textbf{\dname Statistics.} The number of programs, lines of code, and the number of parts are listed. }
    \vspace{-3mm}
    \label{tab:dataset_stats}
    \renewcommand{\arraystretch}{1.2}
    \resizebox{1.0\linewidth }{!}{
    \begin{tabular}{l|c|c|c}
    \toprule[0.4mm]
    \rowcolor{lightgray} 
      & $\#$Programs &
      $\#$Lines (min, median, max) &  $\#$Parts \\ 
      
     \midrule
     
     \dnameCL &1322   &  (21, 40, 66)  & 4,4,4,2   \\
     \dnameCH &1322   &  (61, 72, 162) & 4,4,4,2  \\
     
     \midrule
     
     \dnameEL &1322   &  (7, 115, 1077)  & 4,4,4,2 \\ 
     \dnameEH &1322   &  (27, 162, 1172)  & 4,4,4,2 \\
     
     \midrule
    
     \dnameR & 45   & (28, 120, 381)  & 2-10 \\  
     \bottomrule[0.4mm]
    \end{tabular}
    }
    \vspace{-4mm}
\end{table}

\subsection{Evaluation Metrics}
\label{subsec:metrics}
We propose two metrics to evaluate the performance of algorithms on the new task of commenting CAD programs. The first metric is block accuracy ($B_{acc}$), which calculates the successful rate of block-wise labeling, while the second metric is semantic IoU ($S_{IoU}$), which measures the Intersection-over-Union value per semantic label, averaged over all labels. Detailed formulations of the metrics can be found in the supplemental material.

Intuitively, block accuracy quantifies the general performance, while the semantic IoU considers the label distribution among blocks, making it sensitive to the long-tail problem where some labels only cover a few blocks.

\prag{Evaluation with synonyms.} 
In practice, algorithms addressing the CAD program commenting problem may generate different labels than our ground truth annotations, albeit with a similar semantic meaning. We account for these synonyms in our evaluation by asking ChatGPT-v4 to give us, for a given list of predicted labels, its mapping to the list of ground truth labels (if any). We then apply the mapping before computing the metrics.

\begin{table*}[!htb]
    \centering
    \caption{Evaluation of our method on \dnameC and \dnameE benchmarks. See Table~\ref{tab:comp_partSLIP} for comparison with PartSLIP~\cite{liu2023partslip}.}
    \label{tab:main_stats}
    \vspace{-3mm}
    \renewcommand{\arraystretch}{1.3}
\resizebox{1\textwidth }{!}{
    \begin{tabular}{r|c|c|c|c|c|c|c|c|c|c|c|c|c|c|c|c}
    \toprule[0.4mm]
        
        \multirow{2}*{Dataset}
                     & \multicolumn{8}{c|}{\dnameC} 
                     & \multicolumn{8}{c}{\dnameE} 
                     \\
    
        \cline{2-17}
        
                     & \multicolumn{4}{c|}{\dnameCH} 
                     & \multicolumn{4}{c|}{\dnameCL} 
                     & \multicolumn{4}{c|}{\dnameEH}
                     & \multicolumn{4}{c}{\dnameEL}\\
        \hline
        \rowcolor{lightgray}
            Input Text
                     & \multicolumn{2}{c|}{GPT Words} 
                     & \multicolumn{2}{c|}{GT Words}
                     & \multicolumn{2}{c|}{GPT Words}
                     & \multicolumn{2}{c|}{GT Words} 
                     & \multicolumn{2}{c|}{GPT Words} 
                     & \multicolumn{2}{c|}{GT Words}
                     & \multicolumn{2}{c|}{GPT Words}
                     & \multicolumn{2}{c}{GT Words} \\
        
        \hline
        \hline
        Metric & $B_{acc}$ & $S_{IoU}$ & $B_{acc}$ & $S_{IoU}$ & $B_{acc}$ & $S_{IoU}$ & $B_{acc}$ & $S_{IoU}$ & $B_{acc}$ & $S_{IoU}$ &  $B_{acc}$ & $S_{IoU}$ & $B_{acc}$ & $S_{IoU}$ & $B_{acc}$ & $S_{IoU}$ \\

        \hline 
        
        Airplane (4 parts)
        &\cellcolor{lightgray}   85.03        &     77.76                   
        &\cellcolor{lightgray}  88.00
        & 80.05
        
        &\cellcolor{lightgray}   75.65      
        & 68.40
        &\cellcolor{lightgray} 72.17
        &65.27
        
        &\cellcolor{lightgray}   74.78     
        & 65.77
        &\cellcolor{lightgray} 80.04
        &69.25
        
        &\cellcolor{lightgray}   72.52
        & 65.90
        &\cellcolor{lightgray} 76.60
        & 65.70
        \\
        
        \hline
        Chair (4 parts)
        &\cellcolor{lightgray}   90.02 
        & 84.34                       
        &\cellcolor{lightgray}   88.20
        & 79.42
        
        &\cellcolor{lightgray}   92.28
        &91.11
        &\cellcolor{lightgray} 93.36
        & 90.53
        
        &\cellcolor{lightgray}   77.66     
        &71.29
        
        &\cellcolor{lightgray}81.13
        &73.24
        
        &\cellcolor{lightgray}   71.41
        & 57.95
        &\cellcolor{lightgray} 75.40
        & 61.00
        
        \\
        \hline
        Table (2 parts)
        &\cellcolor{lightgray}  90.88  
        & 86.33                      
        &\cellcolor{lightgray}  90.91
        & 85.64
        
        &\cellcolor{lightgray}   95.76         
        &93.50
        &\cellcolor{lightgray} 90.90
        & 85.89
        
        &\cellcolor{lightgray}  83.06      
        &74.71
        &\cellcolor{lightgray} 83.56
        &74.60
        
        &\cellcolor{lightgray}
        81.21
        &75.76
        &\cellcolor{lightgray} 79.02
        & 72.38
        
        \\
        
        \hline
        Animal (4 parts)
        &\cellcolor{lightgray}  89.07      
        & 82.55
        &\cellcolor{lightgray} 86.15
        & 77.79
        
        &\cellcolor{lightgray}     89.03  
        & 85.68
        
        &\cellcolor{lightgray} 88.70
        & 84.07
        
        &\cellcolor{lightgray}     91.28    
        &83.14
        &\cellcolor{lightgray} 89.42
        & 81.70
        
        &\cellcolor{lightgray}     91.90
        &86.08
        &\cellcolor{lightgray} 85.68
        & 74.81 \\
        
        \hline 
        
        Average & \cellcolor{lightgray} 88.75 & 82.75 & \cellcolor{lightgray} 88.32 & 80.73 & \cellcolor{lightgray} 88.18 & 84.76 & \cellcolor{lightgray} 86.28 & 81.84
        & \cellcolor{lightgray} 81.70 & 73.73 & \cellcolor{lightgray} 83.54 & 74.70 & \cellcolor{lightgray} 79.26 & 71.42 & \cellcolor{lightgray} 79.18 & 68.47 \\
        
        \bottomrule[0.4mm]
    \end{tabular}
    }
\end{table*}

\section{Experiments}
\label{sec:experiments}

We now describe our statistical and visual evaluations, we provide implementation details as supplemental material.

\subsection{Results}
Tab.~\ref{tab:main_stats} reports the quantitative evaluation of our \methodName algorithm on the different tracks of our \dname dataset. 
For each track, we evaluate our algorithm when provided with either the ground-truth list of labels (GT Words) or the list suggested by ChatGPT (GPT Words).
For each setting, we report both the block accuracy and the semantic IoU.
\revi{Our pipeline is not restricted to GPT4. We have also tested with the open-source Llama2-70B (see supplementary).}

Overall, our algorithm achieves similar results in both settings, demonstrating the effectiveness of synonym mapping and the applicability of our algorithm to real-world scenarios where ground truth labels are not available.
The two metrics reach slightly higher values on cuboid than on ellipsoid abstractions, and on high level of details than on low level. Ellipsoid abstractions often contain overlapping primitives, which might be more difficult to label.

The real human-made programs in \dnameR are more diverse in terms of program structure, shape geometry, and shape semantic granularity, leading to a $78.29\%$ and $66.22\%$ for block accuracy and semantic IoU, respectively.
Our method sometimes mislabels parts that are spatially close and semantically related, like the steam dome vs. chimney in Fig.~\ref{fig:teaser}.
Note that the ground truth labels we have specified in our dataset might be biased towards a specific granularity, which influences the metric (e.g., a coarse level of semantics is easier to predict).
Fig.~\ref{fig:vis_main} illustrates typical visual results of our automatic commenting. More results and failure cases can be found in the supplementary.

\subsection{Ablation Study}
\label{subsec:abl_study}

\begin{table*}[!htb]
    \centering
    \caption{Statical evaluation of our ablation study with block accuracy $B_{acc}$.}
    \label{tab:abl_stats}
    \vspace{-3mm}
    
    \renewcommand{\arraystretch}{1.4}
    \resizebox{1\textwidth }{!}{
    \begin{tabular}{c|c|c|c|c|c|c|c|c|c|c|c|c|c|c|c|c}
        \toprule[0.4mm]
                     \multirow{2}*{Dataset}
                     & \multicolumn{8}{c|}{\dnameC} 
                     & \multicolumn{8}{c}{\dnameE} 
                     \\
    
       \cline{2-17}

                     & \multicolumn{4}{c|}{\dnameCH} 
                     & \multicolumn{4}{c|}{\dnameCL} 
                     & \multicolumn{4}{c|}{\dnameEH}
                     & \multicolumn{4}{c}{\dnameEL}\\
        \hline        
        \hline
                     \rowcolor{lightgray}
                     Ab. Cond.
                     & \multicolumn{1}{c|}{w/o MI}
                     & \multicolumn{1}{c|}{w/o CN} 
                     & \multicolumn{1}{c|}{w/o SM}
                     & \multicolumn{1}{c|}{Full}
                     
                     & \multicolumn{1}{c|}{w/o MI}
                     & \multicolumn{1}{c|}{w/o CN} 
                     & \multicolumn{1}{c|}{w/o SM}
                     & \multicolumn{1}{c|}{Full}

                     & \multicolumn{1}{c|}{w/o MI}
                     & \multicolumn{1}{c|}{w/o CN} 
                     & \multicolumn{1}{c|}{w/o SM}
                     & \multicolumn{1}{c|}{Full}

                     & \multicolumn{1}{c|}{w/o MI}
                     & \multicolumn{1}{c|}{w/o CN} 
                     & \multicolumn{1}{c|}{w/o SM}
                     & \multicolumn{1}{c}{Full}
                     
                     \\

        \hline 
        
        Airplane
        & 76.93           
        & 37.02
        & 28.47
        & \textbf{85.03}
        
        & 65.87          
        & 26.41
        & 25.61
        & \textbf{75.65}
        
        &70.91         
        & 25.90 
        &32.84 
        & \textbf{74.78}
        
        & 68.67          
        & 25.92
        &24.83
        &\textbf{72.52}
        \\
        
        \hline
        Chair
        & 73.77           
        & 59.70
        &35.48
        &\textbf{90.02}
        
        & 88.48          
        & 38.50 
        &21.28
        &\textbf{92.28}
        
        & 73.44            
        & 70.62 
        & 33.59
        & \textbf{77.66}
        
        &67.32            
        & 60.93 
        &24.82
        &\textbf{71.41}
        \\
        \hline
        Table
        &   86.35         
        &  79.79
        & 59.90
        & \textbf{90.88}
        
        & 91.95        
        & 80.72 
        &39.23
        &\textbf{95.76}
        
        &  75.44             
        &  80.01
        &50.78
        & \textbf{83.06}
        
        &  77.70          
        &  76.08
        &35.56
        &\textbf{81.21}
        \\
        
        \hline
        Animal
        & 81.75           
        &  25.09
        &56.94
        &\textbf{89.07}
        
        & 74.75           
        & 21.93 
        &38.20
        &\textbf{89.03}
        
        &  86.76          
        & 23.53 
        &48.86
        &\textbf{91.28}
        
        & 87.15           
        &  22.18
        & 49.94
        &\textbf{91.90}
        \\

    \hline
        
        Average & 79.70 & 50.40 & 45.20 & \textbf{88.75} & 80.26 &  41.89 & 31.08 & \textbf{88.18} & 76.64 & 50.02 & 41.52 & \textbf{81.70} & 75.21 & 46.28 & 33.79 & \textbf{79.26}\\
        
        \bottomrule[0.4mm]
    \end{tabular}
    }
    \vspace{-4mm}
\end{table*}

Table~\ref{tab:abl_stats} compares several versions of our algorithm.

\prag{Multi-image generation.} 
Our algorithm generates 4 realistic images for each of the 10 views of the shape. Reducing to one image per view decreases accuracy (Table~\ref{tab:abl_stats}, w/o MI), as the method is more sensitive to ambiguous shading, texture and other artifacts produced by ControlNet~\cite{zhang2023ControlNet}.

\prag{Image-to-image translation.} 
We leverage ControlNet~\cite{zhang2023ControlNet} to turn our renderings of the shape into realistic images. Removing this component dramatically reduces accuracy (Tab.~\ref{tab:abl_stats}, w/o CN) due to the domain gap between our synthetic renderings and the photographs for which Grounded DINO and SAM have been trained.

\prag{Pixel-level segments.} 
We employ an open-vocabulary detection method \cite{liu2023grounding} to predict labeled bounding boxes in images, and we refine these boxes into pixel-level segments using SAM \cite{kirillov2023segment}. 
Tab.~\ref{tab:abl_stats} (w/o SM) shows that accuracy drops significantly if we omit this last segmentation step, and instead rely on box-based IoU to evaluate Eq.~\ref{eq:voting}. Bounding boxes only loosely delineate object parts, resulting in substantial noise in the voting process.

\subsection{Comparison with PartSLIP~\cite{liu2023partslip}}
\label{subsec:comp_ps}

\begin{table}[!tb]
    \centering
    \caption{Statistical Comparison with PartSLIP (PS) and its variants. Block accuracy $B_{acc}$ is reported.}
    \label{tab:comp_partSLIP}
    \vspace{-3mm}
    \renewcommand{\arraystretch}{1.4}
    \resizebox{1.0\linewidth}{!}{
    \begin{tabular}{c|c|c|c|c|c|c|c|c|c|c|c|c}
    \toprule[0.4mm]
        \multirow{2}*{Dataset}
        & \multicolumn{6}{c|}{\dnameC} 
        & \multicolumn{6}{c}{\dnameE}  \\
        
        \cline{2-13}
        
         & \multicolumn{3}{c|}{\dnameCH} 
         & \multicolumn{3}{c|}{\dnameCL} 
         & \multicolumn{3}{c|}{\dnameEH}
         & \multicolumn{3}{c}{\dnameEL} \\
        \hline
        \hline
        \rowcolor{lightgray}
        Methods 
        & \multicolumn{1}{c|}{PS}
        & \multicolumn{1}{c|}{PS++}
        & \multicolumn{1}{c|}{Ours}
        & \multicolumn{1}{c|}{PS}
        & \multicolumn{1}{c|}{PS++}
        & \multicolumn{1}{c|}{Ours}
        & \multicolumn{1}{c|}{PS}
        & \multicolumn{1}{c|}{PS++}
        & \multicolumn{1}{c|}{Ours}
        & \multicolumn{1}{c|}{PS}
        & \multicolumn{1}{c|}{PS++}
        & \multicolumn{1}{c}{Ours} \\
        \hline

        Airplane
        & 22.04
        & 30.07
        & \textbf{85.03}
        & 14.51
        & 23.31
        & \textbf{75.65}
        & 22.84
        & 23.88
        & \textbf{74.78}
        & 17.83
        & 20.61
        & \textbf{72.52} \\
        
        \hline
        
        Chair
        & 42.04
        & 51.01
        & \textbf{90.02}
        & 36.82
        & 44.90
        & \textbf{92.28}
        & 37.97
        & 46.40
        & \textbf{77.66}
        & 43.00
        & 49.12
        & \textbf{71.41} \\

        \hline
        
        Table
        & 17.83
        & 56.79
        & \textbf{90.88}
        & 16.23
        & 58.07
        & \textbf{95.76}
        & 20.55
        & 61.96
        & \textbf{83.06}
        & 17.56
        & 58.01
        & \textbf{81.21} \\

        \hline
        
        Animal
        & 36.76
        & 55.64
        & \textbf{89.07}
        & 38.26
        & 54.40
        & \textbf{88.18}
        & 23.33
        & 42.32
        & \textbf{81.70}
        & 18.48
        & 35.48 
        & \textbf{91.90} \\

        \hline
        
        Average
        & 29.67
        & 48.38
        & \textbf{88.75}
        & 26.46
        & 45.17
        & \textbf{87.97}
        & 26.17
        & 43.64
        & \textbf{79.30}
        & 24.22
        & 40.80
        & \textbf{79.26} \\

        \bottomrule[0.4mm]
    \end{tabular}
    }
    \vspace{-6mm}
\end{table}

Given that CAD programs are another type of 3D representation, our semantic program-commenting task can be considered a zero-shot, open-set 3D part segmentation problem. We compare our method with the state-of-the-art zero-shot 3D point cloud segmentation method PartSLIP~\cite{liu2023partslip}. 

Specifically, we convert \dname shapes into point clouds and feed them to PartSLIP. Since PartSLIP outputs point-wise labels, we aggregate the point labels back to the program blocks based on block-point correspondence. Each block is assigned a label corresponding to most of the points it contains. Table~\ref{tab:comp_partSLIP} displays the statistical comparison, where the block accuracy is much lower than ours.
We hypothesize that this performance drop is because PartSLIP relies on traditional rendering of the point cloud to perform visual reasoning, which results in non-realistic images in our context.
Besides, PartSLIP assigns a null label to points that cannot be semantically recognized. Since null labels decrease our accuracy metrics, we also implemented a version that assigns a random label drawn from the list of ground-truth labels to each code block not recognized by PartSLIP (Tab.~\ref{tab:comp_partSLIP}, PS++). While this variant achieves higher metric values, it remains inferior to ours. The same observation applies to \dnameR (e.g., $78.29\%$ vs. $18.06\%$ for ours and PS in terms of $B_{acc}$); our prediction is better. Please refer to the supplemental.

\subsection{Commenting on Machine-made Macros}
\label{subsec:shapecoder}

While the machine-made programs in our dataset exhibit a flat structure, methods like ShapeCoder~\cite{jones2023shapecoder} can automatically discover abstractions within flat programs to form libraries of nested functions. We provide as supplemental materials an experiment on ShapeCoder programs, showing that while \methodName produces comments that convey the semantic meaning of the functions, these functions sometimes mix several semantic parts since ShapeCoder created them solely based on geometry.

\subsection{Semantic Commenting using ChatGPT}
\label{subsec:LLM}

The key idea of \methodName is to execute and render the CAD shape to cast program commenting as an image segmentation task. While our evaluation on \dname demonstrates the effectiveness of this image-based strategy, it 
can struggle in the presence of small parts and occlusions.

These limitations motivated us to experiment with a program-based strategy, for which visibility and appearance are irrelevant.
Specifically, inspired by recent successes of few-shot training of LLMs for code commenting \cite{Brown2020,Ahmed2023}, as well as by ongoing effort on leveraging LLMs to help designers in various tasks \cite{makatura2023large}, we instructed ChatGPT-v4 to comment on a program given another program with comments as an example.
We provide the results of this experiment as supplemental materials, which reveals that ChatGPT succeeds in commenting on programs that are similar to the example (same object category, same geometric primitives) but fails to generalize to new shapes and primitives.

\section{Conclusion}
In this paper, we have introduced the new problem of semantic commenting of CAD programs and proposed a novel method, \methodName, to tackle this problem by combining program parsing with visual-semantic analysis offered by recent advances in foundational language and vision models. We have established an effective baseline for this new problem. Additionally, we have prepared a new benchmark dataset -- \dname~-- to evaluate the performance of our algorithm and to facilitate future research in this direction.

\prag{Limitations.} Our current algorithm cannot perform any non-trivial reordering or restructuring of input CAD programs. Hence, if an input program is badly written, instead of simply being badly commented, we cannot improve it to make it more readable. This leaves out an interesting optimization axis as even the exact same geometric shape can have many different programmatic realizations. 
Our method can only comment on instructions that produce geometry, not on instructions that remove geometry (which we aggregate within irreducible blocks). For example, if a shape is subtracted from another shape to form a hole, we cannot comment on the semantic meaning of that hole (as for the window of the cab of the train in Fig~\ref{fig:teaser}). \revi{A possible solution would be to explicitly render the geometry removed by the subtraction (i.e., for $A-B$, render $A\cap B$).}
Finally, the \dnameR track is currently limited to 45 models, which is too little to serve as training data.

\prag{Future work.} 
Our early experiments with ChatGPT (Sec~\ref{subsec:LLM}) suggest that a multi-modality approach (e.g., taking as input tuples of programs, 2D renderings, and 3D shapes) is a promising direction for future work as it would allow reasoning on both the program structure and its visual output.
Also, our ability to populate flat machine-generated programs with comments could benefit analysis of these programs, for instance, to account for both the geometry and semantics of the shape when searching for program abstractions \cite{jones2021shapemod,jones2023shapecoder}.
Finally, we would also like to assign meaningful labels to the program parameters to make it easier for human users to edit the programs (e.g., renaming the variable that controls how tall a chair back is to `height'). Vision models capable of reasoning about \emph{differences} between images \cite{Park_2019_ICCV} or semantic 3D features~\cite{dutt2024diffusion} might help. 

\revi{\prag{Acknowledgments.} 
We thank Algot Runeman for his \href{http://runeman.org/3d/}{OpenSCAD programs}. CJ was supported by a startup grant from the School of Informatics, Bayes Seed funding, and gifts from Google Cloud research credits. NM was supported by Marie Skłodowska-Curie grant agreement No. 956585 and UCL AI Centre.}

{
\small
\bibliographystyle{ieeenat_fullname}
\bibliography{main}
}

\clearpage
\setcounter{page}{1}
\maketitlesupplementary

\section{Additional Results}
\subsection{Commenting on ShapeCoder~\cite{jones2023shapecoder} Programs}

\begin{figure*}[!htb]
    \centering
    \begin{overpic}[width=\linewidth]{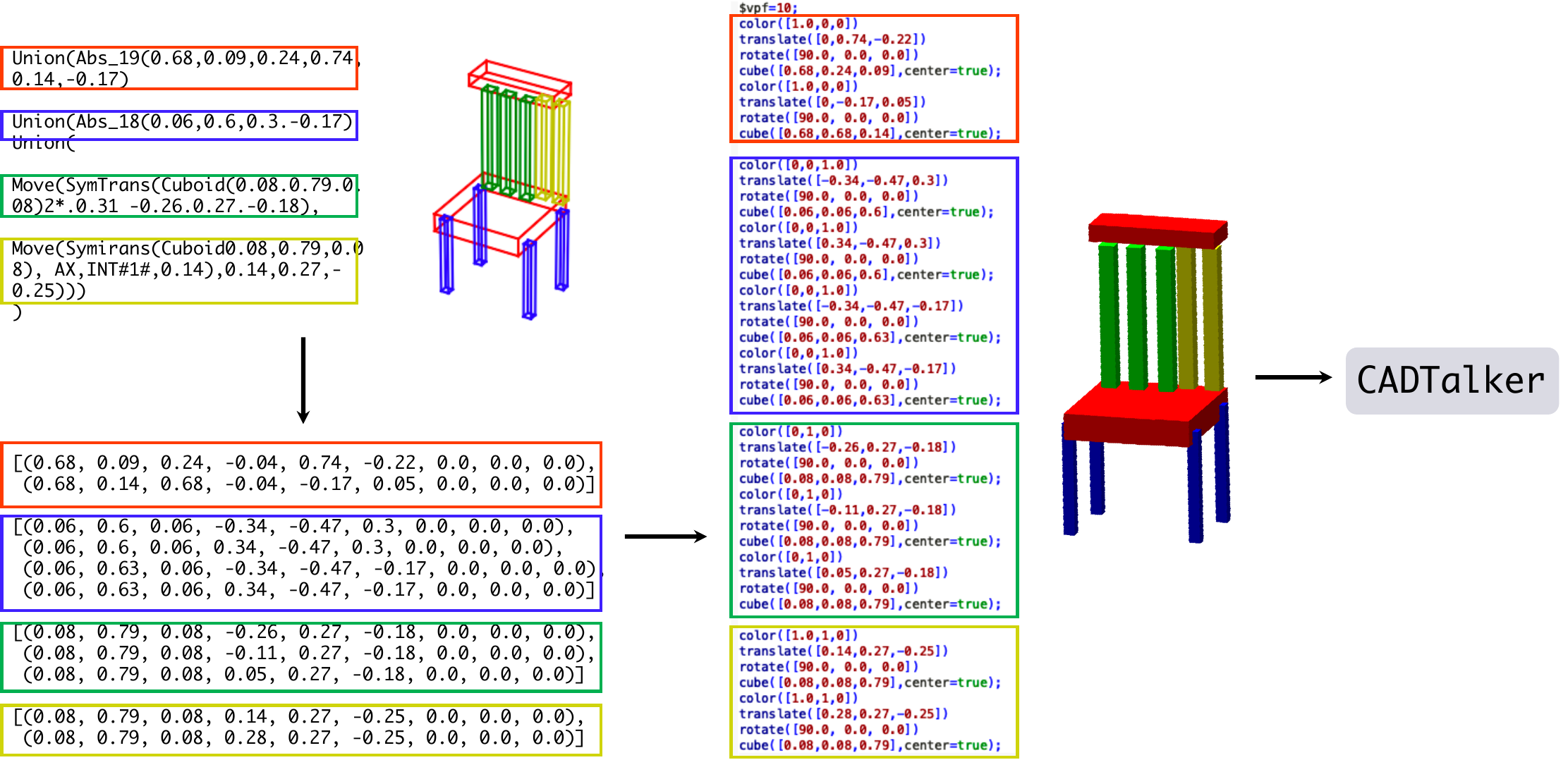} 
        \put(25, 44) {\small (a)}
        \put(36, 21) {\small (b)}
        \put(68, 38) {\small (c)}
        \put(86, 27) {\small (d)}
    \end{overpic}
    \caption{\textbf{ShapeCoder Program Processing}. Given a ShapeCoder program (a), we first execute the program to obtain all primitive parameters, i.e., a set of cubes represented by width, height, length, rotation, and translation (b). Then, we translate these primitives into an OpenSCAD program (c) and run our \methodName (d). The colorful boxes indicate the line-parameter-code block correspondence.}
    \label{fig:shapecoder-process}
\end{figure*}


\begin{figure*}[!htb]
\centering
\begin{minipage}[t]{0.68\textwidth}
\centering
\includegraphics[width=\textwidth]{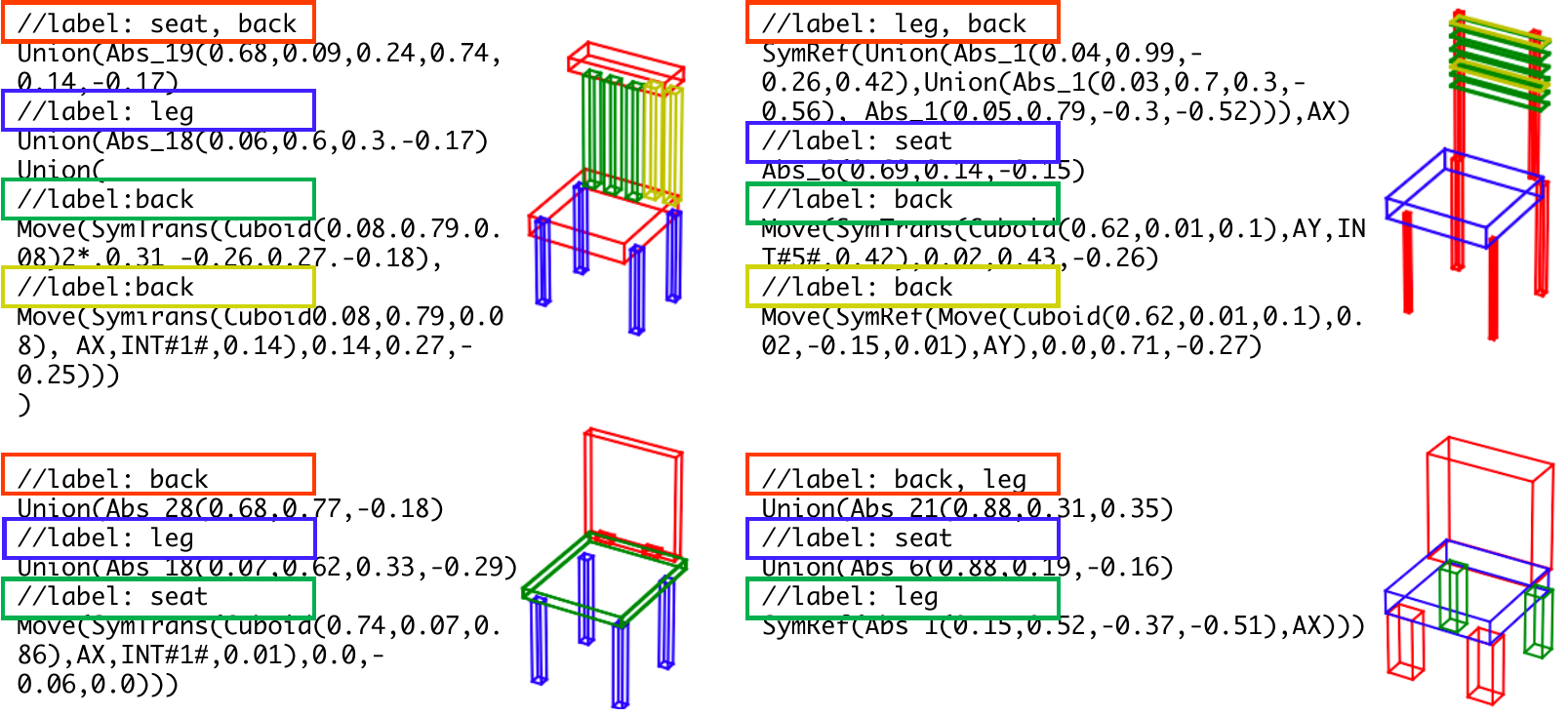}
(a)
\end{minipage}
\hfill
\begin{minipage}[t]{0.285\textwidth}
\centering
\includegraphics[width=\textwidth]{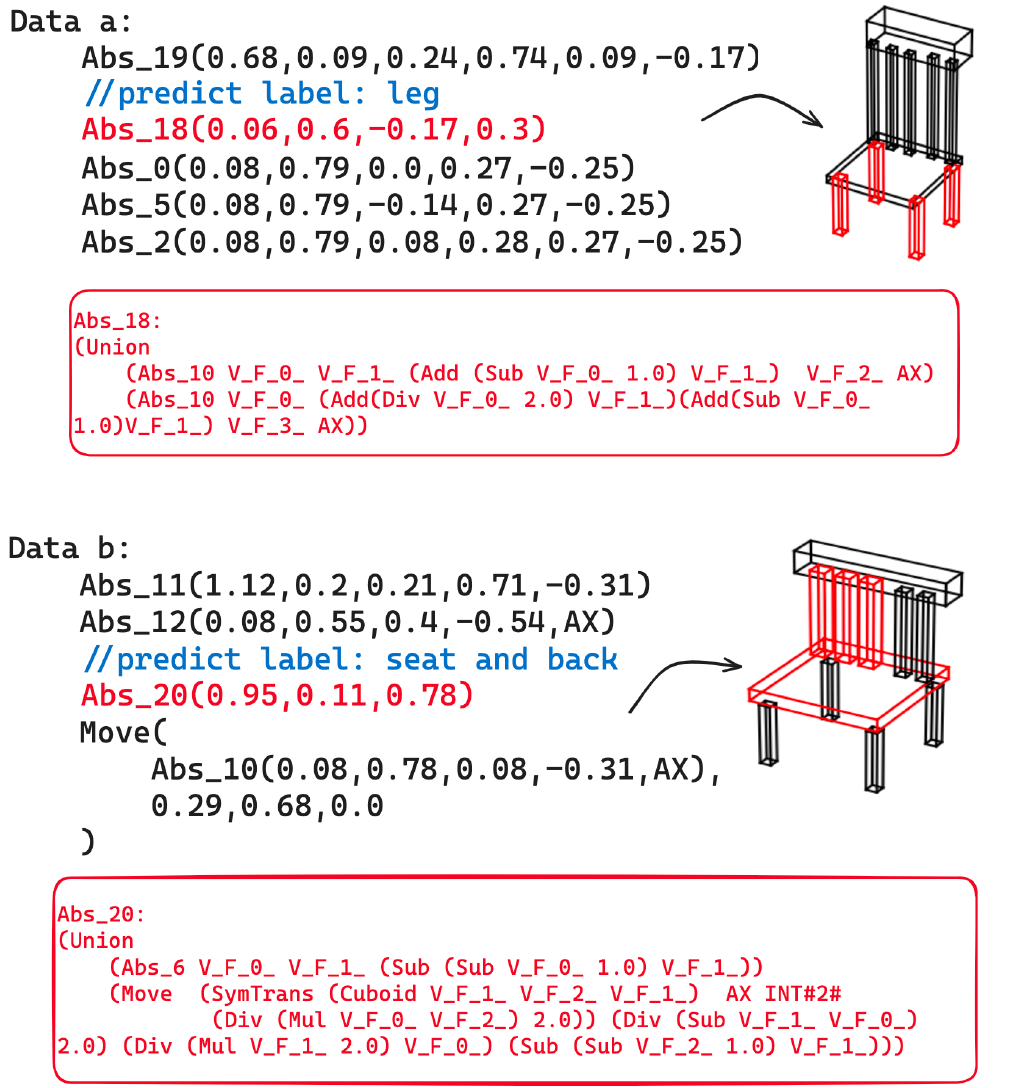}
(b)
\end{minipage}
\vspace{-2mm}
\caption{(a) Four typical commenting results on ShapeCoder programs with colored boxes indicating the comment-shape correspondence. (b) ShapeCoder~\cite{jones2023shapecoder} identifies redundancy in shape datasets to generate code macros (red blocks) that encapsulate common parts. While our approach produces descriptive comments for these macros (blue comments), the macros themselves do not always correspond to isolated semantic parts (bottom blue comments). }
\label{fig:ShapeCoder}
\end{figure*}

While the machine-made programs in our dataset exhibit a flat structure, methods like ShapeCoder~\cite{jones2023shapecoder} can automatically discover abstractions within flat programs to form libraries of nested functions.
We have tested \methodName on the abstracted shape programs from ShapeCoder. 

\prag{Data processing and running.} Since ShapeCoder only provides a simple executor that produces line renderings (Fig.~\ref{fig:shapecoder-process} (a)), we resort to OpenSCAD as an alternative executor to obtain 3D shapes suitable for depth map rendering. 
Specifically, as shown in Fig.~\ref{fig:shapecoder-process}, for each ShapeCoder program, we first use its default executor to transform the program into cuboid primitives represented by a set of parameters (e.g., height, width, and translation).
We then translate these cuboid primitives into an OpenSCAD program, which can be executed to get the required depth map. 
After the translation, each line of the ShapeCoder program corresponds to one or more OpenSCAD code lines.
We run \methodName to generate the semantic comment for each line and aggregate these comments for each ShapeCoder line by a simple non-repetitive merging.
We transfer the semantic comments back to the ShapeCoder program by exploiting the recorded program line and code block correspondence.

\prag{Results.} Fig.~\ref{fig:ShapeCoder} (a) shows typical results of our algorithm on programs produced by ShapeCoder. While our comments convey the semantic meaning of the ShapeCoder functions, they also reveal that because the ShapeCoder algorithm solely works on geometry, it produces functions that mix semantic parts (Fig.~\ref{fig:ShapeCoder} (b)). This experiment suggests that automatic commenting could serve as a way to evaluate the semantic coherence of automatically generated code macros.

\subsection{Semantic Commenting using ChatGPT}

\begin{table}[!tb]
\begin{center}
\caption{Different configurations when interacting with GPT-4 to comment on a cuboid airplane program. Each row (b-e) corresponds to a different commented example provided to GPT-4 for one-shot training.}
\label{tab:gpt_config}
\scalebox{1.0}{
\begin{tabular}{c|c|l|c}
\toprule[0.4mm]
    \rowcolor{lightgray}
    Option & Teach  & Example Program & $B_{acc}$$ (\uparrow)$ \\ \midrule
    a & \xmark  & \xmark & $31.88$ \\
    b & \cmark &  Airplane$^{Cube}$ (partial) & $52.5$ \\
    c & \cmark & Airplane$^{Ellip}$ & $37.5$ \\
    \rowcolor{lightgray}
    d & \cmark &  Airplane$^{Cube}$ & $\mathbf{87.5}$ \\
    e & \cmark & Chair$^{Cube}$ & $44.37$ \\
\bottomrule[0.4mm]
\end{tabular}
}
\end{center}
\end{table}

\begin{figure}[!tb]
    \centering
    \includegraphics[width=\linewidth]{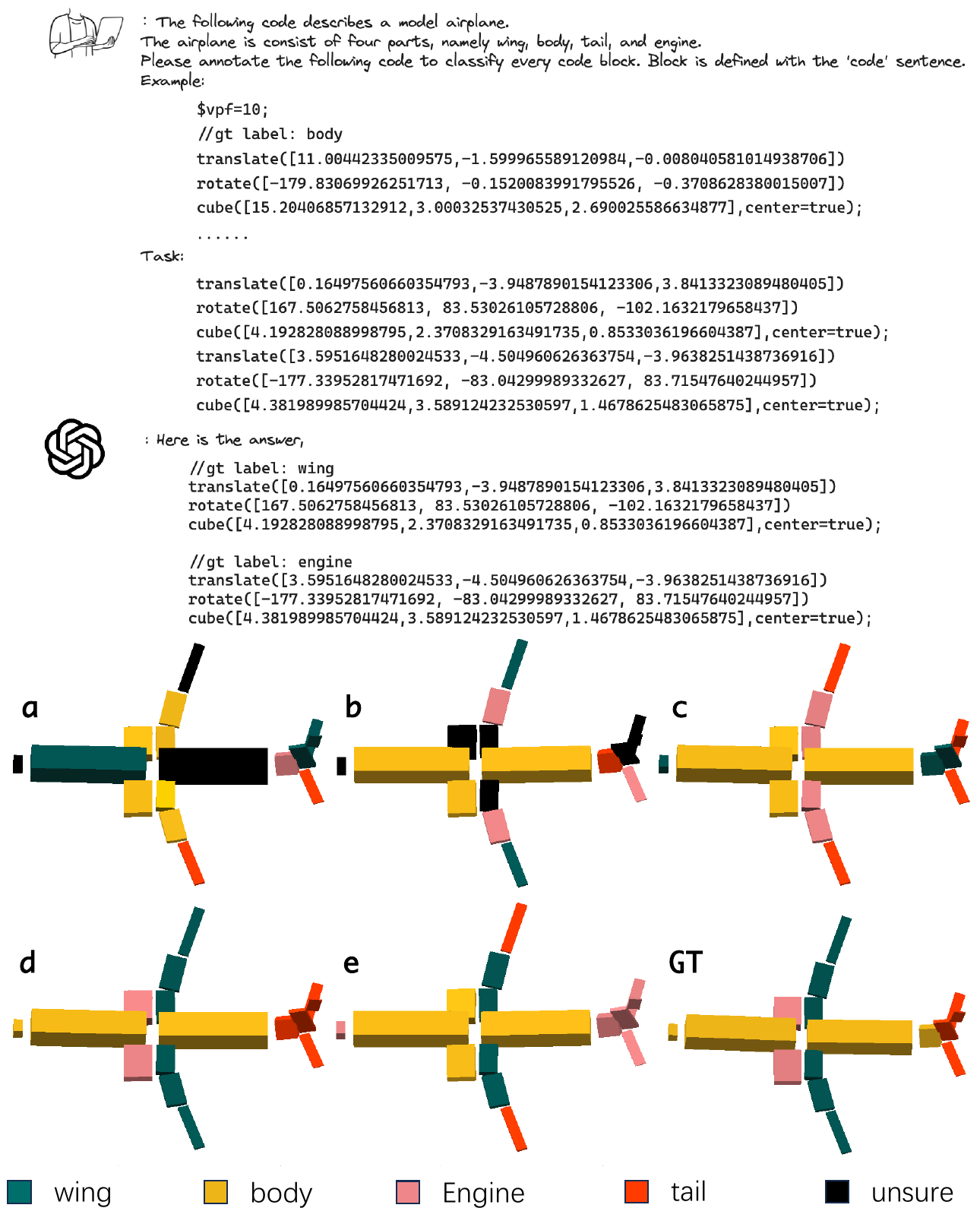}
    \vspace{-6mm}
    \caption{ChatGPT commenting results under different configurations.}
    \label{fig:gpt_seg}
\end{figure}

The key idea of the algorithm we have proposed -- \methodName -- is to execute and render the CAD shape to cast program commenting as an image segmentation task. While our evaluation on \dname demonstrates the effectiveness of this image-based strategy, it has some limitations.
First, object parts can be occluded in most of the views, and as such do not get labeled. Similarly, small parts that only cover a few pixels tend to be ignored. Second, while our use of image-to-image translation greatly reduces the domain gap between renderings and photographs, the images we obtain might still contain unrealistic details that are difficult to recognize.

These limitations motivated us to also experiment with a program-based strategy, for which visibility and appearance are irrelevant.
Specifically, inspired by recent successes of few-shot training of LLMs for code commenting \cite{Brown2020,Ahmed2023}, we instructed ChatGPT-v4 to comment on an airplane program from our \dnameC dataset.
In addition to the program to be commented on, we also provided ChatGPT with the list of part names (i.e., 'body', 'wings', 'tail', and 'engine'), as illustrated in Fig.~\ref{fig:gpt_seg}, top row.

Within this setup, we tested five different configurations of the commenting task, as listed Tab.~\ref{tab:gpt_config} where the superscript indicates the source of the example program.
\begin{itemize}
    \item Option (a): zero-shot prediction, \emph{no} additional information is provided to ChatGPT.
    \item Option (b): one-shot prediction, the example is \emph{incomplete} and comes from the same \emph{airplane} category in \dnameC.
    \item Option (c): one-shot prediction, the example is \emph{complete} but made of \emph{different} primitives (i.e., ellipsoid) from \dnameE.
    \item option (d): one-shot prediction, the example is \emph{complete} and comes from the same \emph{airplane} category in \dnameC.
    \item option (e): one-shot prediction, the example is \emph{complete} but from the \emph{chair} category in \dnameC.
\end{itemize}
In all cases, we shuffle the code blocks of both the task program and the example program to avoid any influence of ordering. This experiment reveals that, to our surprise, a single example is enough for ChatGPT to successfully comment programs that represent the same object category, with the same geometric primitives (configuration d). 
When asked to explain its answer, ChatGPT reported using the volume and relative position of the parts as evidence, such as the fact that the body should have the biggest volume, while the wings should be attached on the two sides of the body \footnote{GPT is trained to produce the next word given the prompt, we are not sure if it effectively used volume and positions to solve the task. It is an interesting research direction to reveal the mechanisms of GPT.}. However, the other configurations (b,c,e) reveal that these spatial reasoning skills do not extend to examples that are incomplete, of another category, or made of different primitives. A small-scale statistical evaluation of these configurations with 10 testing examples is reported in Tab.~\ref{tab:gpt_config}, which is consistent with the above analysis and the visual results in Fig.~\ref{fig:gpt_seg}, where the configuration (d) achieves the best result. 

The full conversations with ChatGPT-v4 can be found in a separate file titled ``GPT-Conversation.pdf'' \revi{on the project page}.

\subsection{Comparison with PartSLIP}

\begin{table*}[!htb]
    \centering
    \caption{Statistical Comparison with PartSLIP and its variants.}
    \label{tab:comp_PS_full}
    \vspace{-3mm}
    \renewcommand{\arraystretch}{1.6}
    \resizebox{1.0\textwidth}{!}{
    \begin{tabular}{c|c|c|c|c|c|c|c|c|c|c|c|c|c|c|c|c|c|c|c|c|c|c|c|c}
    \toprule[0.4mm]
        \multirow{2}*{Dataset}
        & \multicolumn{12}{c|}{\dnameC} 
        & \multicolumn{12}{c}{\dnameE}  \\
        
        \cline{2-25}
        
         & \multicolumn{6}{c|}{\dnameCH} 
         & \multicolumn{6}{c|}{\dnameCL} 
         & \multicolumn{6}{c|}{\dnameEH}
         & \multicolumn{6}{c}{\dnameEL} \\
        \hline
        \hline
        \rowcolor{lightgray}
        Methods 
        & \multicolumn{2}{c|}{PS}
        & \multicolumn{2}{c|}{PS++}
        & \multicolumn{2}{c|}{Ours}
        & \multicolumn{2}{c|}{PS}
        & \multicolumn{2}{c|}{PS++}
        & \multicolumn{2}{c|}{Ours}
        & \multicolumn{2}{c|}{PS}
        & \multicolumn{2}{c|}{PS++}
        & \multicolumn{2}{c|}{Ours}
        & \multicolumn{2}{c|}{PS}
        & \multicolumn{2}{c|}{PS++}
        & \multicolumn{2}{c}{Ours} \\
        \hline
        Metric & $B_{acc}$ & $S_{IoU}$ & $B_{acc}$ & $S_{IoU}$ & $B_{acc}$ & $S_{IoU}$ & $B_{acc}$ & $S_{IoU}$ & $B_{acc}$ & $S_{IoU}$ &  $B_{acc}$ & $S_{IoU}$ & $B_{acc}$ & $S_{IoU}$ & $B_{acc}$ & $S_{IoU}$ & $B_{acc}$ & $S_{IoU}$ & $B_{acc}$ & $S_{IoU}$ & $B_{acc}$ & $S_{IoU}$ & $B_{acc}$ & $S_{IoU}$\\
        \hline
        Chair
        & \cellcolor{lightgray} 62.81 & 59.74
        & \cellcolor{lightgray} 64.51 & 59.35
        & \cellcolor{lightgray} \textbf{90.02} & \textbf{84.34}
        & \cellcolor{lightgray} 74.24 & 60.29
        & \cellcolor{lightgray} 74.90 & 59.60
        & \cellcolor{lightgray} \textbf{92.28} & \textbf{91.11} 
        & \cellcolor{lightgray} 64.92& 57.46
            & \cellcolor{lightgray} 64.94& 57.20
        & \cellcolor{lightgray} \textbf{77.66} & \textbf{71.29}
        & \cellcolor{lightgray} 66.86& 52.23 
        & \cellcolor{lightgray} 67.54 & 51.99
        & \cellcolor{lightgray} \textbf{71.41} & \textbf{57.95} \\
        
        \hline
        
        Airplane
        & \cellcolor{lightgray} 21.94 & 9.76
        & \cellcolor{lightgray} 29.36 &11.94
        & \cellcolor{lightgray} \textbf{85.03} & \textbf{77.76}
        & \cellcolor{lightgray} 14.40 &8.01 
        & \cellcolor{lightgray} 22.71 & 12.24
        & \cellcolor{lightgray} \textbf{75.65} & \textbf{68.40}
        & \cellcolor{lightgray} 23.16 & 8.95 
        & \cellcolor{lightgray} 25.15 & 9.84
        & \cellcolor{lightgray} \textbf{74.78} & \textbf{65.77}
        & \cellcolor{lightgray} 19.74 & 8.56
        & \cellcolor{lightgray} 23.99 & 10.40
        & \cellcolor{lightgray} \textbf{72.52} & \textbf{69.90} \\

        \hline
        
        Animal
        & \cellcolor{lightgray} 41.94 & 32.26 
        & \cellcolor{lightgray} 59.07 & 34.65
        & \cellcolor{lightgray} \textbf{89.07} & \textbf{82.75 }
        & \cellcolor{lightgray} 44.69 &  41.70
        & \cellcolor{lightgray} 55.55 & 44.47
        & \cellcolor{lightgray} \textbf{89.03} & \textbf{85.68}
        & \cellcolor{lightgray} 33.96 & 25.54 
        & \cellcolor{lightgray} 44.12 & 29.71
        & \cellcolor{lightgray} \textbf{91.28} & \textbf{83.14}
        & \cellcolor{lightgray} 29.96 & 24.37
        & \cellcolor{lightgray} 42.94 & 25.86
        & \cellcolor{lightgray} \textbf{91.90} &\textbf{86.08}  \\

        \hline
        
        Table
        & \cellcolor{lightgray} 40.90 & 33.43
        & \cellcolor{lightgray} 64.98 & 49.64
        & \cellcolor{lightgray} \textbf{90.88} & \textbf{86.33}
        & \cellcolor{lightgray} 37.60 &  27.55
        & \cellcolor{lightgray} 65.61 & 49.50
        & \cellcolor{lightgray} \textbf{95.76} & \textbf{93.50} 
        & \cellcolor{lightgray} 38.38 & 35.02
        & \cellcolor{lightgray} 63.34 & 45.09
        & \cellcolor{lightgray} \textbf{83.06} & \textbf{74.71}
        & \cellcolor{lightgray} 43.05 & 32.04
        & \cellcolor{lightgray} 61.38 & 53.66
        & \cellcolor{lightgray} \textbf{81.21} & \textbf{75.76} \\

        \hline
        
        Average 
        &\cellcolor{lightgray}41.90& 33.80
        & \cellcolor{lightgray}54.48& 38.89
        &\cellcolor{lightgray} \textbf{88.75}& \textbf{82.80}
        &\cellcolor{lightgray} 42.73& 34.39
        &\cellcolor{lightgray} 54.69& 41.45
        &\cellcolor{lightgray} \textbf{88.18}& \textbf{84.67}
        &\cellcolor{lightgray} 40.10& 31.74
        &\cellcolor{lightgray} 49.39& 35.46
        &\cellcolor{lightgray} \textbf{81.69}& \textbf{73.73}
        &\cellcolor{lightgray} 39.9& 29.30
        &\cellcolor{lightgray} 48.96& 35.48
        &\cellcolor{lightgray} \textbf{79.26}& \textbf{72.42} 
        \\
        
        \bottomrule[0.4mm]
    \end{tabular}
    }

\end{table*}

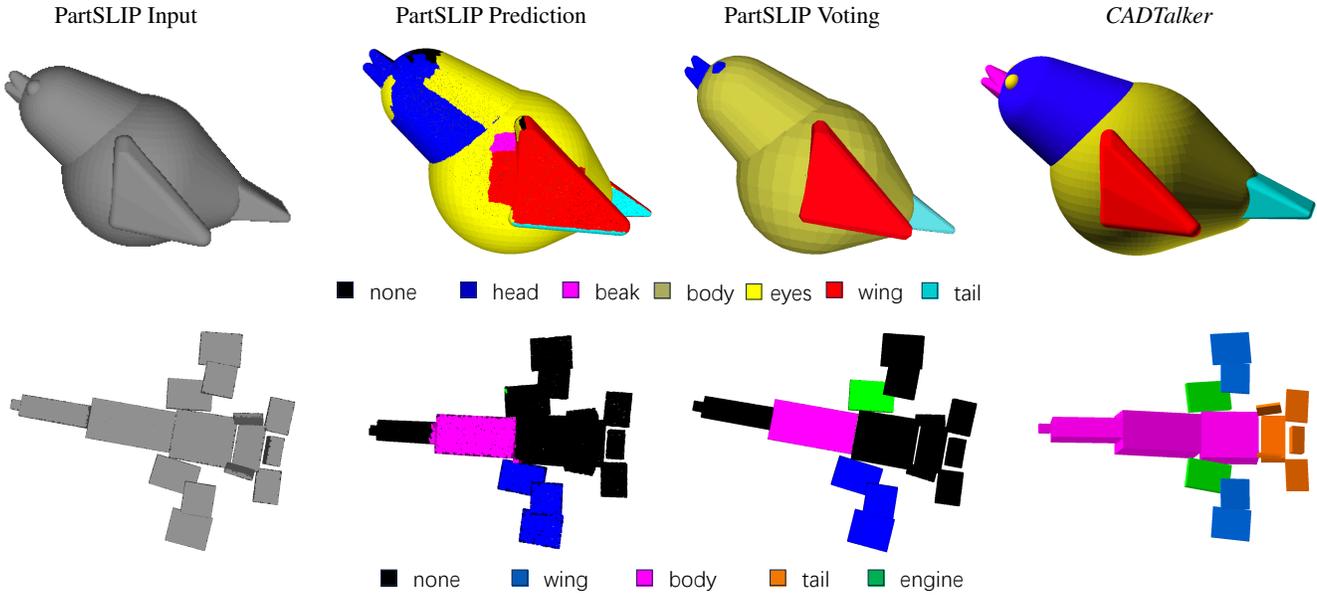
\begin{figure*}[!tb]
    \centering
    \begin{overpic}[width=\textwidth]{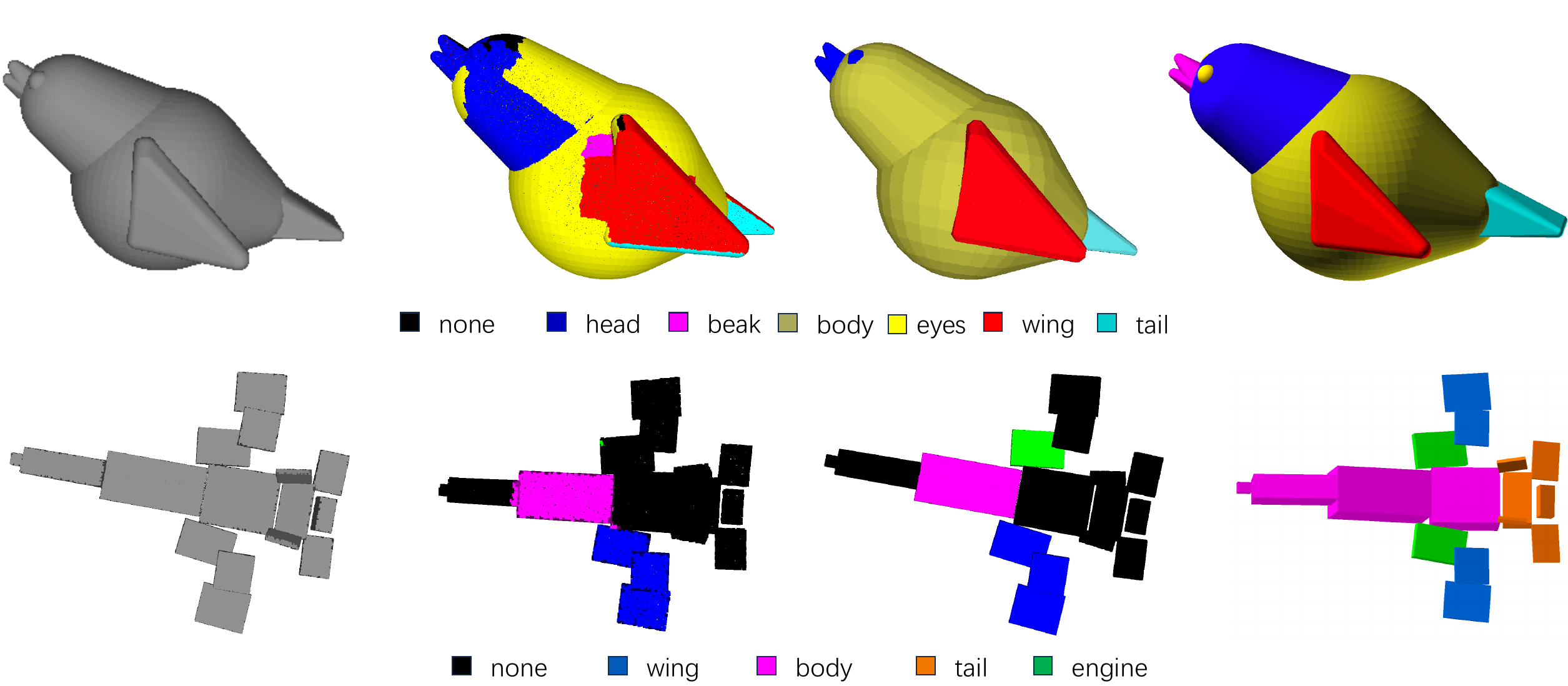} 
        \put(4, 43.5) {\small PartSLIP Input}
        \put(30, 43.5) {\small PartSLIP Prediction}
        \put(55, 43.5) {\small PartSLIP Voting}
        \put(84, 43.5) {\small \methodName}
    \end{overpic}
    \vspace{-6mm}
    \caption{\textbf{Visual Comparison with PartSLIP}.
    Due to the absence of realistic colors, the raw prediction of the per-point label is noisy, leaning toward missing many points (the black color). After the label aggregation, errors are still obvious, e.g., the tail and most of the body of the airplane are mislabeled, while the head, beak, and eyes of the bird are totally wrong.
    }
    \label{fig:comp_partslip}
\end{figure*}

In \cref{subsec:comp_ps}, we have described the comparison with PartSLIP regarding block accuracy. Here, we elaborate on the details and provide more statistical and visual results. 

\prag{Data processing and running.} Taking as input a dense and colored 3D point cloud and part names as prompts, PartSLIP predicts point-wise labels belonging to the part names. 
To compare, we first execute each program from \dname to obtain a 3D model and densely sample it with 200K points with normal and a uniform gray color (see Fig.~\ref{fig:comp_partslip}, PartSLIP Input). As for the point-based rendering, we implemented a simple Phong shading\footnote{The original PartSLIP code for point-based rendering does not apply any shading because it assumes that the input point cloud is a 3D colored scan. We replaced that code with our simple Phong shading. Some numbers reported in Tab.~\ref{tab:comp_partSLIP} of our submission were computed with the original rendering, which resulted in a lower performance. Nevertheless, even with better shading, PartSLIP's results are far inferior to ours. We will revise the numbers upon acceptance.} to produce the images fed to PartSLIP.
In the sampling process, we record the point-block correspondence for label transferring using a similar binary mask based registration procedure as described in \cref{subsec:label_voting}.
This resulting point cloud is fed into PartSLIP to obtain point-wise labels. We then aggregate the point-wise labels of each commentable block by choosing the label with the highest number of votes and simply transfer the resulting label back to the shape program as the predicted comments.

\prag{Results.} Full statistics with both the block accuracy ($B_{acc}$) and the semantic IoU ($S_{IoU}$) are shown in Tab.~\ref{tab:comp_PS_full}, where we obtain far better results compared with PartSLIP and PartSLIP++. As for the human-made program, the block accuracy for PartSLIP, PartSLIP++, and ours are $38.17\%$ vs. $39.24\%$ vs. $78.29\%$, while the semantic IoUs are $27.25\%$, and $27.73\%$, and $66.22\%$, respectively. Visual Results can be found in Fig.~\ref{fig:comp_partslip}.
PartSLIP fails this zero-shot point cloud segmentation task on both machine-made and human-made programs in our context. This is mainly attributed to PartSLIP's strong dependency on point clouds that incorporate realistic colors, a feature frequently absent in program representations. 
This failure is further evidenced in our ablation study, wherein the exclusion of ControlNet (w/o CN) results in notably reduced evaluation metrics.

\subsection{Comparison with SATR}
\begin{figure}[!tb]
    \centering
    \includegraphics[width=\linewidth]{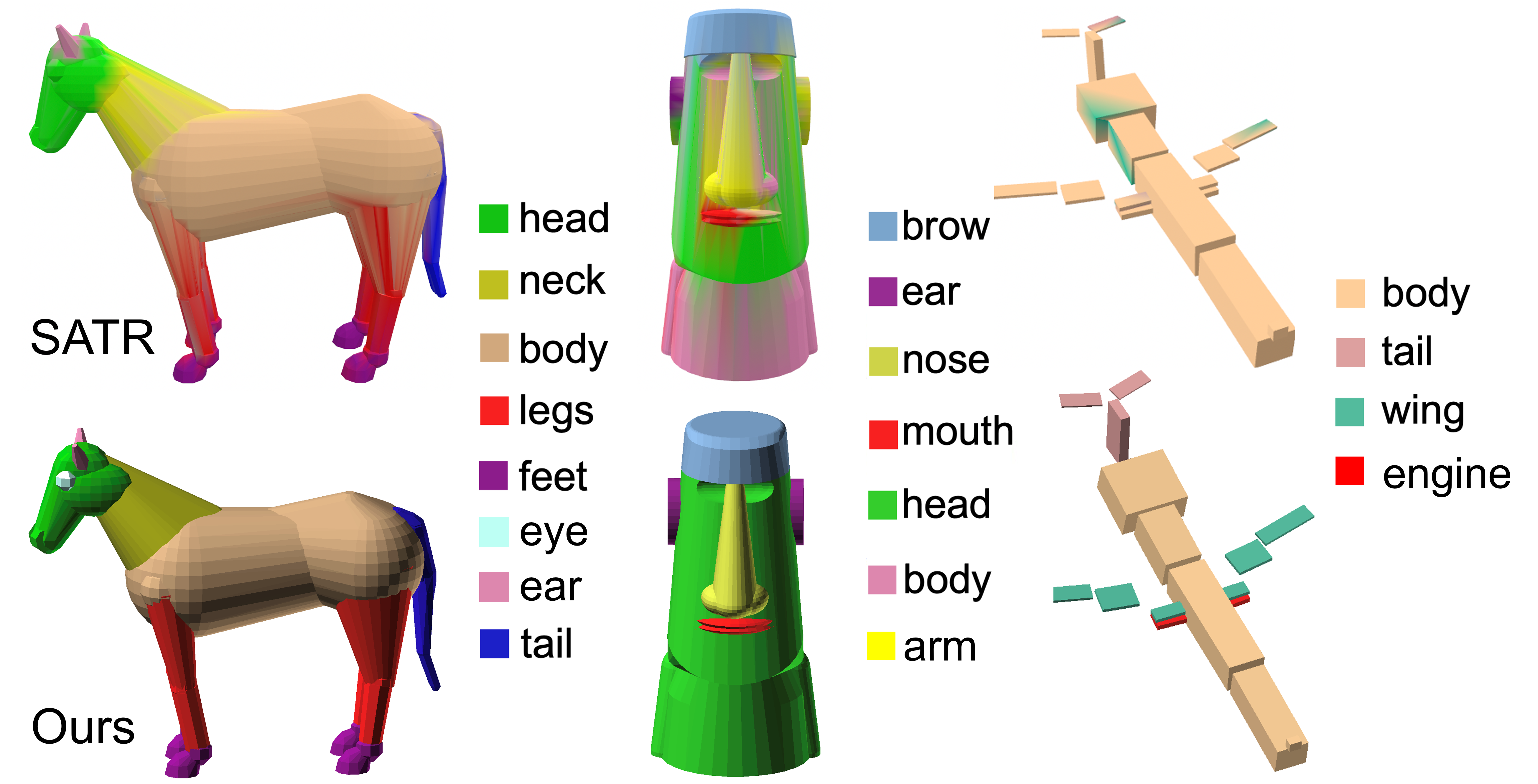}
    \vspace{-6mm}
    \caption{\textbf{Visual Comparison with SATR}. The first row shows the results from SATR, while our results are in the second row.
    }
    \vspace{-4mm}
    \label{fig:comp_satr}
\end{figure}

\revi{As discussed in \cref{ses:rw} and \cref{subsec:comp_ps}, our task can be considered as a zero-shot, open-set 3D part segmentation problem. Other than PartSLIP, we preliminarily compare our method with SATR \cite{abdelreheem2023satr}, the state-of-the-art zero-shot 3D mesh segmentation method. Qualitative results are shown in Fig. \ref{fig:comp_satr}, where mesh segmentations are competitive on the realistic horse, but SATR struggles on the more abstracted Moai sculpture and fails on the abstracted airplane. The reason is the gap between the rendered images from abstracted shapes and the photographs used for training the large image-language model, while ControlNet in our pipeline solves this problem effectively.
}

\subsection{OpenLLM Model Test}
\revi{Our pipeline is highly modular and not restricted to GPT4, we thus test our algorithm with the open-source Llama2-70B model. Statistical results are displayed in Table \ref{tab:llama2_test}, where performance degradation is observed. For example, with Llama2-70B, the $B_{acc}$ is dropped from $88.75\%$ to $82.96\%$ and $S_{IoU}$ is dropped from $82.75\%$ to $75.43\%$ on \dnameCH programs.
As for the real human-made programs in CADTalk-Real, Llama2-70B achieves $70.88\%$ and $57.97\%$ for block accuracy and semantic IoU, which are reduced by $7.4\%$ and $8.3\%$ compared with GPT-4 (i.e., $78.29\%$ and $66.22\%$, respectively).
Once a more powerful LLM is available, our method can enjoy the improvement without any special tunning.
}

\begin{table*}[!htb]
    \centering
    \caption{Comparison between GPT words and LLAMA2 words on the full dataset.}
    \label{tab:llama2_test}
    \vspace{-3mm}
    \renewcommand{\arraystretch}{1.3}
\resizebox{1\textwidth }{!}{
    \begin{tabular}{r|c|c|c|c|c|c|c|c|c|c|c|c|c|c|c|c}
    \toprule[0.4mm]
        
        \multirow{2}*{Dataset}
                     & \multicolumn{8}{c|}{\dnameC} 
                     & \multicolumn{8}{c}{\dnameE} 
                     \\
    
        \cline{2-17}
        
                     & \multicolumn{4}{c|}{\dnameCH} 
                     & \multicolumn{4}{c|}{\dnameCL} 
                     & \multicolumn{4}{c|}{\dnameEH}
                     & \multicolumn{4}{c}{\dnameEL}\\
        \hline
        \rowcolor{lightgray}
            Input Text
                     & \multicolumn{2}{c|}{GPT Words} 
                     & \multicolumn{2}{c|}{LLAMA2 Words}
                     & \multicolumn{2}{c|}{GPT Words}
                     & \multicolumn{2}{c|}{LLAMA2 Words} 
                     & \multicolumn{2}{c|}{GPT Words} 
                     & \multicolumn{2}{c|}{LLAMA2 Words}
                     & \multicolumn{2}{c|}{GPT Words}
                     & \multicolumn{2}{c}{LLAMA2 Words} \\
        
        \hline
        \hline
        Metric & $B_{acc}$ & $S_{IoU}$ & $B_{acc}$ & $S_{IoU}$ & $B_{acc}$ & $S_{IoU}$ & $B_{acc}$ & $S_{IoU}$ & $B_{acc}$ & $S_{IoU}$ &  $B_{acc}$ & $S_{IoU}$ & $B_{acc}$ & $S_{IoU}$ & $B_{acc}$ & $S_{IoU}$ \\

        \hline 
        
        Airplane (4 parts)
        &\cellcolor{lightgray}   85.03        &     77.76                   
        &\cellcolor{lightgray}  85.71
        &  78.28
        
        &\cellcolor{lightgray}   75.65      
        & 68.40
        &\cellcolor{lightgray} 77.32
        & 71.51
        
        &\cellcolor{lightgray}   74.78     
        & 65.77
        &\cellcolor{lightgray} 77.43
        &71.19
        
        &\cellcolor{lightgray}   72.52
        & 65.90
        &\cellcolor{lightgray} 75.90
        & 70.66
        \\
        
        \hline
        Chair (4 parts)
        &\cellcolor{lightgray}   90.02 
        & 84.34                       
        &\cellcolor{lightgray}   87.18
        & 77.33
        
        &\cellcolor{lightgray}   92.28
        &91.11
        &\cellcolor{lightgray} 91.47
        & 88.11
        
        &\cellcolor{lightgray}   77.66     
        &71.29
        
        &\cellcolor{lightgray}75.49
        &68.56
        
        &\cellcolor{lightgray}   71.41
        & 57.95
        &\cellcolor{lightgray} 68.99
        & 55.30
        
        \\
        \hline
        Table (2 parts)
        &\cellcolor{lightgray}  90.88  
        & 86.33                      
        &\cellcolor{lightgray}  80.71
        & 73.47
        
        &\cellcolor{lightgray}   95.76         
        &93.50
        &\cellcolor{lightgray} 88.19
        & 79.96
        
        &\cellcolor{lightgray}  83.06      
        &74.71
        &\cellcolor{lightgray} 74.35
        &61.27
        
        &\cellcolor{lightgray}
        81.21
        &75.76
        &\cellcolor{lightgray} 78.23
        & 67.11
        
        \\
        
        \hline
        Animal (4 parts)
        &\cellcolor{lightgray}  89.07      
        & 82.55
        &\cellcolor{lightgray} 78.26
        & 72.65
        
        &\cellcolor{lightgray}     89.03  
        & 85.68
        
        &\cellcolor{lightgray} 86.82
        & 86.46
        
        &\cellcolor{lightgray}     91.28    
        &83.14
        &\cellcolor{lightgray} 75.19
        & 74.85
        
        &\cellcolor{lightgray}     91.90
        &86.08
        &\cellcolor{lightgray} 71.05
        & 70.29 \\
        
        \hline 
        
        Average & \cellcolor{lightgray} 88.75 & 82.75 & \cellcolor{lightgray} 82.96 & 75.43 & \cellcolor{lightgray} 88.18 & 84.76 & \cellcolor{lightgray} 85.95 & 81.51
        & \cellcolor{lightgray} 81.70 & 73.73 & \cellcolor{lightgray} 75.62 & 68.97 & \cellcolor{lightgray} 79.26 & 71.42 & \cellcolor{lightgray} 74.54 & 65.84 \\
        
        \bottomrule[0.4mm]
    \end{tabular}
    }
\end{table*}

\subsection{Additional Commenting Results}
Typical commenting results can be seen on the accompanying webpage with highlighted code and block animations, and more commenting results from all data tracks in \dname can be found in a separate file titled ``Commenting-Results.pdf'' \revi{on the project page}.
In the following, we introduce typical failure cases. 

\begin{figure*}[!htb]
    \begin{overpic}[width=\textwidth]{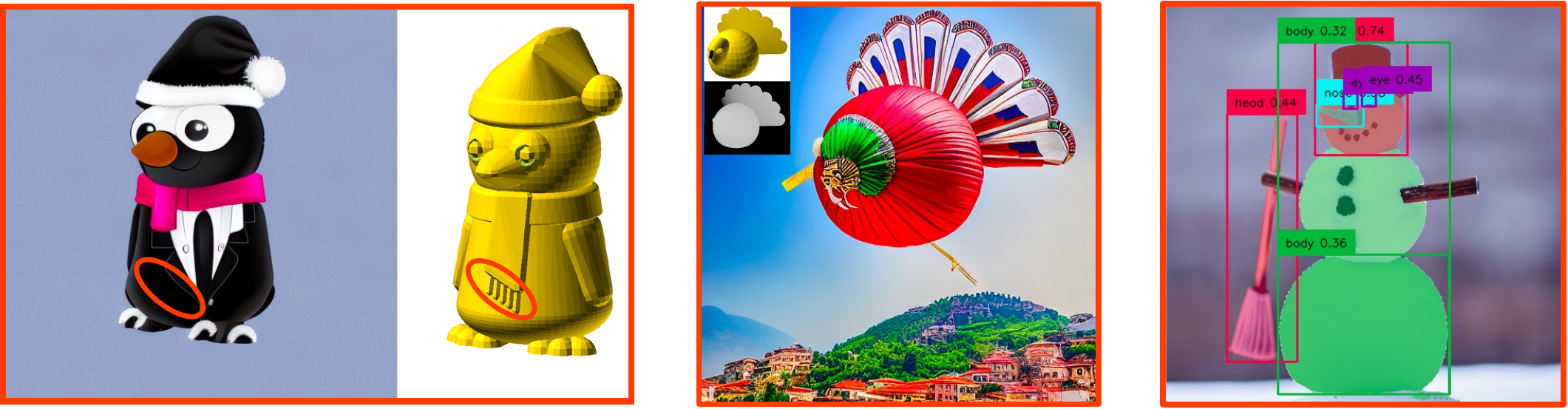} 
        \put(19, -1.6) {\small (a)}
        \put(56, -1.6) {\small (b)}
        \put(86, -1.6) {\small (c)}
    \end{overpic}
    \caption{\textbf{Failure Cases}. (a) ControlNet fails to generate scarf tassels. (b) ControlNet generates an unexpected image given a turkey depth map and keyword, as if it confused ``turkey'' (bird) with ``Turkey'' (country). (c) Grounding DINO wrongly predicts the broom to be `head'.
    }
    \label{fig:failure}
\end{figure*}

\prag{Failure cases.} In Fig.~\ref{fig:failure}, we illustrate typical failure cases of our method, which are mainly inherited from foundational vision-language models, i.e., ControlNet may ignore fine details of the input depth map or generate totally unrecognizable images, and Grounding DINO may mislabel parts that can be seen clearly in the image. 

To address these issues, potential solutions include a) utilizing stronger vision-language models with enhanced conditional generation ability, and more robust detection ability and b) implementing an image discriminator to exclude problematic images, which we leave for future work.

\section{Method Details}

\subsection{Implementation Details}
For depth map processing, we use morphological closing \cite{haralick1987image} with varied configurations. Specifically, we apply 5 iterations of closing for the abstract shapes of \dnameCH and \dnameEH, 3 iterations for \dnameCL and \dnameEL, and 1 iteration for \dnameR, using a $3\times3$ structuring element. 
When using ControlNet~\cite{zhang2023ControlNet}, we set the control strength to 1.0, DDIM sampling steps to 20, the image instance number to 4, and the image resolution to $512\times 512$. We use a simple text prompt template -- ``[CateName], realistic" for ControlNet, where [CateName] is the category name, e.g., Chair. 
We employ default parameter configurations from Grounding DINO~\cite{liu2023grounding} and SAM~\cite{kirillov2023segment} without additional adjustments or tuning.
For our voting scheme, we render depth maps from 10 viewpoints that are evenly distributed around a circular path centering on the object's up axis and maintaining an elevation angle of 55 degrees above the object. When filling in the cumulative confidence score, we progressively adjust the filtering threshold in the aforementioned three steps (\cref{subsec:label_voting}), setting it at 0.001, 0.01, and 0.02, respectively.

\prag{Running Time}. All experiments were conducted with a single RTX3090 GPU. \revi{Using our unoptimized code, for a program with 200 lines, the overall running time is around 6mins, distributed as $0.2\%$ for program parsing, $1.1\%$ for depth images rendering, $85.1\%$ for ControlNet, $0.7\%$ for prompt querying, $12.5\%$ for DINO+SAM and $0.3\%$ for voting. }

\subsection{Program Parsing}
\begin{figure*}[!htb]
    \centering
    \includegraphics[width=0.62\textwidth]{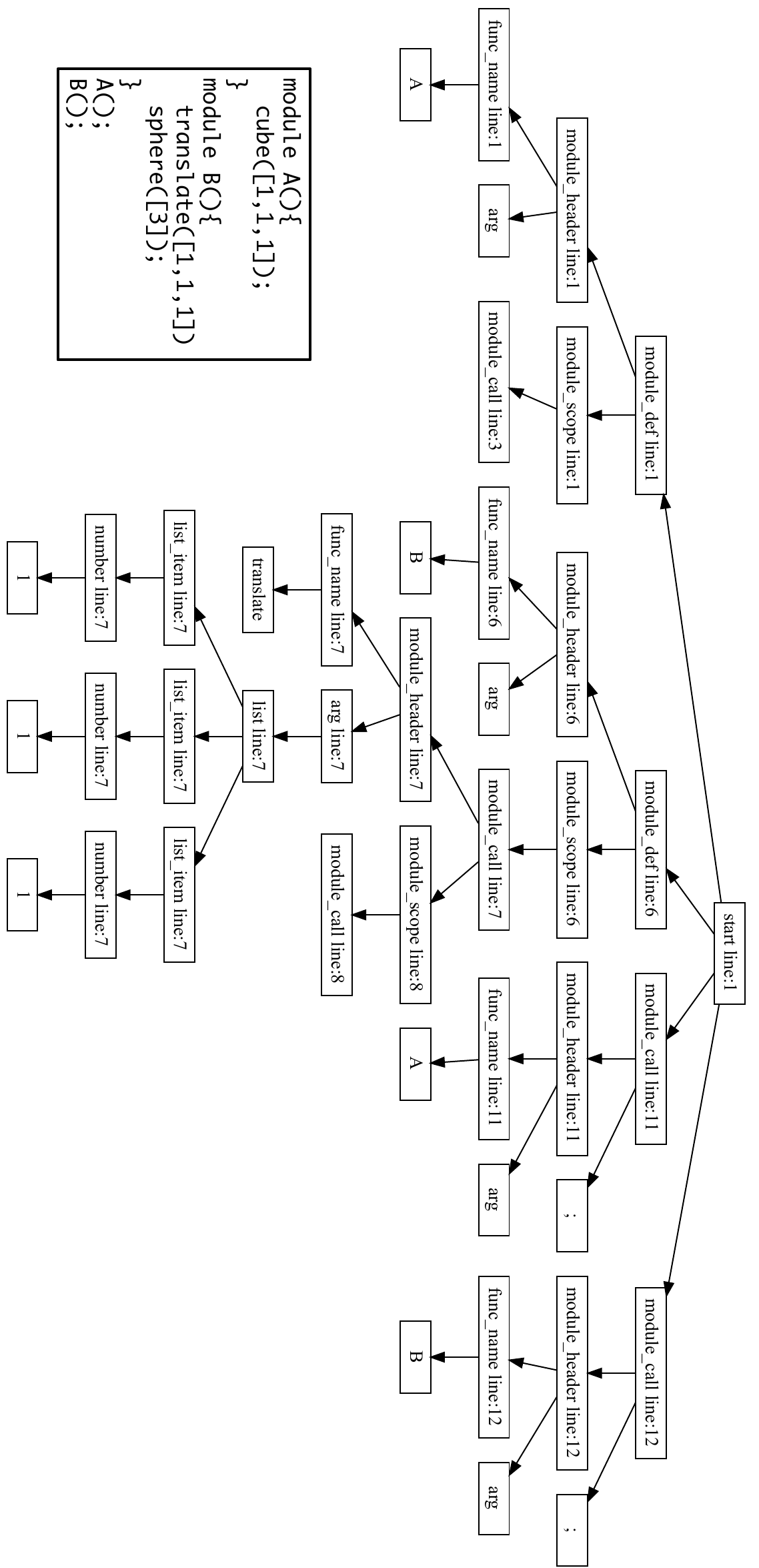}
    \caption{Abstracted Syntax Tree (AST). Each node in the AST maintains the operation type and the corresponding line number for pixel-block registration.}
    \label{fig:parsing}
\end{figure*}

In ~\cref{subsec:label_voting}, we introduced program parsing that produces an Abstract Syntax Tree (AST), laying the foundation for our commenting task. 
To do this, we exploit Lark~\cite{Lark} to conduct lexical and syntax analysis following the OpenSCAD grammar, and the analysis procedure generates an analysis tree, which is equivalent to the original program and wherein all the operation information is stored in the node and the code structure is maintained in the tree structure.
Then, we construct the AST by traversing the analysis tree, and choose the required information, i.e., node type and line number, from the tree node. 
Example AST of a simple program is shown in Fig.~\ref{fig:parsing}, and more trees of programs in \dname can be found in the file titled ``AST.pdf'' \revi{on the project page}. 

\section{\dname Dataset}

\subsection{Dataset Overview}
To facilitate evaluation and foster future research on the semantic CAD program commenting task, we have introduced a new benchmark -- \dname, a dataset of OpenSCAD programs enriched with part-based semantic comments. Tab.~\ref{tab:dataset_stats_detailed} shows detailed statistics per category of each data track, including the number of code lines, and the number of parts.

\begin{figure}[!htb]
    \centering
    \includegraphics[width=\linewidth]{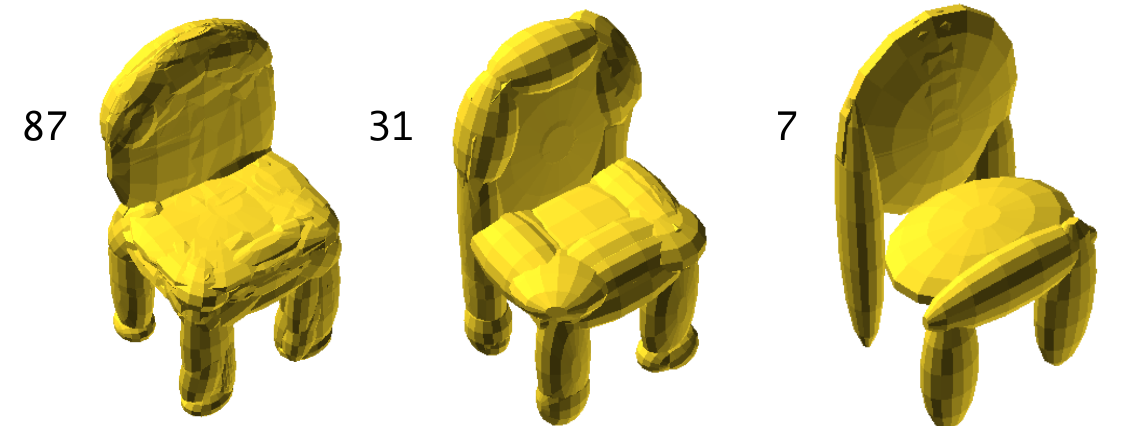}
    \vspace{-6mm}
    \caption{\textbf{Shape abstraction levels.} A chair in \dnameE with different numbers of ellipsoids.}
    \label{fig:abs_level}
\end{figure}

We considered two distinct sources of programs, i.e., human-made and machine-made programs, in our dataset. Since it is difficult to find and manually comment on real shape programs, we only gathered 45 such programs with rich shape and program diversity and we plan to keep collecting more in the future.
For machine-made programs, we rely on automatic methods that convert 3D shapes into cuboid~\cite{sun2019learning} and ellipsoid~\cite{liu2023marching}. One feature of this data track is the two levels of details of the programs, where the ones with a high level of detail reconstruct the shape well but have more lines to comment on, while the others with a low level of detail are harder to recognize due to the abstraction. See Fig.~\ref{fig:abs_level} for an example.

\begin{table}[!htb]
    \centering
    \caption{\textbf{Detailed \dname Statistics.} The number of programs, lines of code, and the number of parts per category for each data track are listed.}
    \vspace{-3mm}
    \label{tab:dataset_stats_detailed}
    \renewcommand{\arraystretch}{1.3}
    \resizebox{1.0\linewidth }{!}{
    \begin{tabular}{l|c|c|c|c}
    \toprule[0.4mm]
    \rowcolor{lightgray} &Category
      & $\#$Programs &
      $\#$Lines (min, median, max) &  $\#$Parts \\ 
      
     \midrule
     
     \multirow{4}*{\dnameCL} &airplane &400   &  (40, 40, 40)  & 4  \\
                             &chair &400   &  (66, 66, 66)  & 4  \\
                             &table &400   &  (21, 21, 21)  & 2  \\
                             &animal &122   &  (40, 40, 40)  & 4   \\

     \midrule

     \multirow{4}*{\dnameCH} &airplane &400  &  (72, 72, 72) & 4  \\
                             &chair &400  &  (162, 162, 162) & 4  \\
                             &table &400   &  (61, 61,61 ) & 2  \\
                             &animal &122   &  (72, 72, 72) & 4  \\
     
     \midrule
     
     \multirow{4}*{\dnameEL} &airplane &400   &  (37, 100, 242)  & 4\\ 
                             &chair &400   &  (27, 147, 672)  & 4 \\ 
                             &table &400   &  (7, 101, 1077)  & 2 \\ 
                             &animal &122   &  (62, 112, 166)  & 4 \\ 
     \midrule

     \multirow{4}*{\dnameEH} &airplane &400   &  (32, 163, 237)  & 4 \\
                             &chair &400   &  (57, 261, 842)  & 4 \\
                             &table &400   &  (27, 178, 1172)  & 2 \\
                             &animal &122   &  (27, 152, 245)  & 4 \\

     \midrule
    
     \dnameR &real & 45   & (28, 120, 381)  & 2-10 \\  
     \bottomrule[0.4mm]
    \end{tabular}
    }
    \vspace{-4mm}
\end{table}

\subsection{Evaluation Metrics}
We have proposed two metrics to evaluate the performance of algorithms on the new task of commenting CAD programs. In the following, we introduce the formulations to calculate them.
\begin{itemize}
    \item \emph{Block accuracy} is the block-wise labeling accuracy, defined as:
\begin{equation}
        B_{acc} = \frac{m}{n},
        \label{eq:b_acc}
\end{equation}
where $m$ counts the number of blocks that get the correct label and $n$ is the total number of blocks.
\item \emph{Semantic IoU} measures the Intersection-over-Union value per semantic label, averaged over all labels:
\begin{equation}
    S_{IoU} = \frac{1}{K} \sum_{k} \frac{\{l_k\} \cap \{l_k^*\}}{\{l_k\} \cup \{l_k^*\}},
    \label{eq:g_acc}
\end{equation}
where K is the number of labels, $\{l_k\}$ is the set of code blocks predicted to be of the $k^{th}$ label, $\{l_k^*\}$ is the set of code blocks with the $k^{th}$ label as ground truth.
\end{itemize}

\subsection{Machine-made Program Processing}
Given machine-generated shape primitives of ShapeNet models, we turn them into OpenSCAD programs and then conduct automatic labeling and manual refinement.

\prag{Program Translation.} Given the cube or ellipsoid primitives represented by corresponding parameters, we trivially translate these primitives into OpenSCAD cube or ellipsoid primitives, following the same procedure as described in Fig.~\ref{fig:shapecoder-process}. 
Specifically, we translate a cube represented by its eight corners into the native cube primitive in OpenSCAD, while we translate an ellipsoid presented by its semi-axe lengths, rotation, and translation parameters into the native ellipsoid primitive in OpenSCAD.

\prag{Automatic Labels Transferring.}
Since cubes or ellipsoids are generated based on 3D models from PartNet, the existing part labels in PartNet can be utilized for part label assignment. 
Specifically, given a shape program, we first convert the corresponding PartNet shape into a point cloud with per-point labels. We then compare the part shape generated by each code block to the labeled point cloud by checking the IoU, and obtain the corresponding part label by maximum voting.
For a part, e.g., the airplane engine, it may occupy both the wing and engine areas, we thus keep all valid labels in the voting.

\begin{figure}[!tb]
    \centering
    \includegraphics[width=0.9\linewidth]{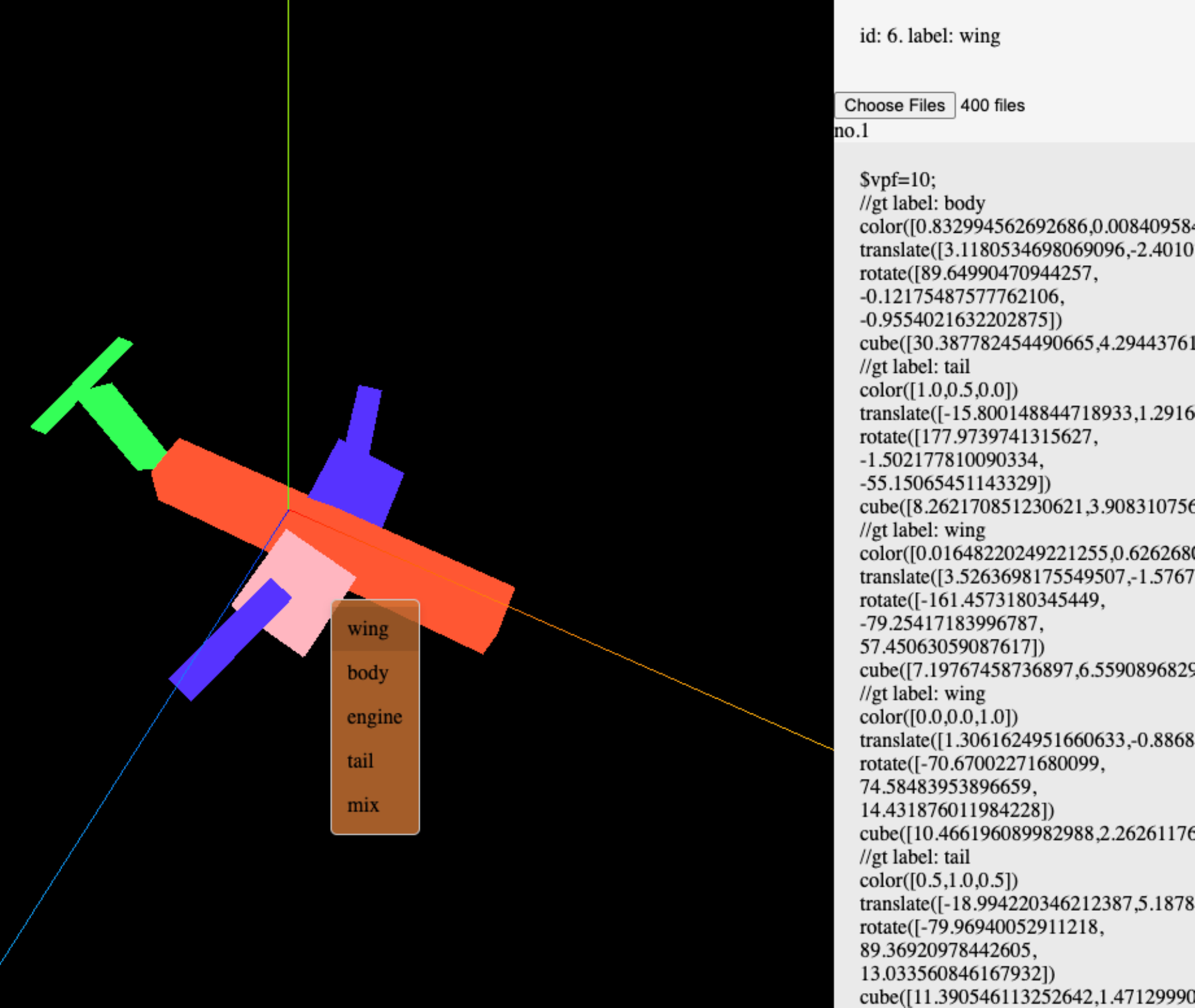}
    \caption{\textbf{User Interface.} The interface enables users to efficiently go through programs and adjust labels.}
    \label{fig:ui}
\end{figure}

\prag{Label Refinement with a Developed UI.} For further refinement of the automatically generated labels, we also developed an interactive UI~(Fig.~\ref{fig:ui}) to directly review and adjust labeled programs in \dname by simple mouse clicking and keyboard hitting.

\end{document}